%% file: main.tex
\documentclass{article}

\usepackage{PRIMEarxiv}
\usepackage[utf8]{inputenc}
\usepackage[T1]{fontenc}
\usepackage{hyperref}
\usepackage{url}
\usepackage{booktabs}
\usepackage{amsfonts}
\usepackage{nicefrac}
\usepackage{microtype}
\usepackage{graphicx}
\graphicspath{{../figures/}}
\usepackage{amsmath}
\usepackage{multirow,makecell,adjustbox,comment,svg} 
\usepackage{fancyhdr}
\usepackage{cleveref}

\title{From Physics to Machine Learning and Back:\\
Part II - Learning and Observational Bias in PHM}
\author{
    \normalfont
  Olga Fink\thanks{Corresponding author: olga.fink@epfl.ch} , Ismail Nejjar\thanks{Equal Contributions}, Vinay Sharma$^\dagger$, Keivan Faghih Niresi, Han Sun, Hao Dong, Chenghao Xu, \\ Amaury Wei, Arthur Bizzi, Raffael Theiler, Yuan Tian, Leandro Von Krannichfeldt, Zhan Ma,\\  Sergei Garmaev, Zepeng Zhang, Mengjie Zhao
  \\
  \\
  Intelligent Maintenance and Operations Systems Lab, EPFL, Lausanne, Switzerland 
}
\date{\today}

\begin{document}
\maketitle

\begin{abstract}
\input{PaperII/00_abstract}
\end{abstract}

\keywords{Prognostics and Health Management \and Physics‑Informed Machine Learning \and Learning Bias \and Observational Bias \and PINNs \and Virtual Sensing \and Data Augmentation \and Reinforcement Learning \and Domain Generalization}

\section{Introduction}
\input{PaperII/01_Introduction}

\section{Leveraging Learning Bias: Loss Functions and Constraints for Robust PHM Models}\label{learning_bias}
\subsection{Physics-Informed Neural Networks (PINNs)}
\input{PaperII/02_MODEL_PINN}
\label{PINNS}

\subsection{Degradation Dynamics Informed Prognostics Methods}
\input{PaperII/04_SOFT_PHYS}
\label{sec:soft_phys}

\section{Leveraging Observational Biases: Augmentation and Physical Constraints}\label{observational_bias}

\subsection{Soft Sensing}

\input{PaperII/05_METHODS_SOFTSENSING}
\label{SS}

\subsection{Overcoming the Simulation-to-Real Gap}
\input{PaperII/06_Simulation_to_real_gap}

\label{sim2real}

\subsection{Generative Modeling}
\input{PaperII/07_METHOD_GENERATIVE_MODEL}
\label{sec:generative}

\subsection{Data Fusion}
\input{PaperII/12_METHOD_DATA_FUSION}
\label{sec:data_fusion}

\section{Reinforcement Learning}

\input{PaperII/08_METHOD_REINFORCEMENT_LEARNING}
\label{RL}

\section{Scaling Beyond Single Systems} \label{scaling}

Physics-informed machine learning improves over purely data-driven methods by increasing model robustness and encouraging physically-consistent behavior. However, PHM methods still face deployment and scalability challenges, particularly in fleet-level scenarios.

When a model trained on data from a single unit is applied to other units within the same fleet, performance often degrades significantly \cite{wang2019domain}. Indeed, operational conditions can vary across assets, with many scenarios not captures in the original training data \cite{michauFleetPHMCritical2018, huPrognosticsHealthManagement2022}. For example, in fault diagnosis, a well-trained deep model may underperform on a new machine simply due to unseen environmental or operational conditions. Similarly, if a machine’s conditions evolve over time, the model may misinterpret these changes as faults if they deviate too much from the training distribution.

These challenges arise from the observational bias inherent in real-world data: models are rarely trained on datasets that capture the full range of operating conditions across a fleet,  even though all units may share the same underlying physics \cite{chengKnowledgeTransferAdaptive2024}. This issue is further exacerbated by the limited availability of labeled data, as industrial datasets are often sparsely annotated due to the high cost of labeling.

While domain adaptation techniques aim to address distributional gaps between source and target domains, they typically assume access to sufficient unlabeled target data. However, in practice, PHM applications often involve small, imbalanced, and task-specific datasets with limited or no labels\textemdash conditions under which conventional learning methods perform poorly
\cite{fengMetalearningPromisingApproach2022}.

To overcome these challenges, fast adaptation and domain generalization have emerged as promising strategies for scaling PHM beyond single-system training. These techniques aim to enable quick model specialization or improve robustness across diverse conditions:

\begin{itemize}
    \item \textbf{Fast Adaptation Methods} (Section \ref{sec:few_shot}): These techniques enable models to quickly specialize to new assets using only a small number of labeled examples, either by learning optimal starting parameters that can be fine-tuned with minimal data, or by leveraging in-context learning approaches that adapt without parameter updates.

     \item \textbf{Domain generalization methods} (Section \ref{gen}): These approaches aim to learn invariant representations that remain stable across diverse operating conditions.
\end{itemize}

\subsection{Fast Adaptation Methods}
\input{PaperII/10_METHOD_FAST_ADAPTATION}
\label{sec:few_shot}

\subsection{Domain Generalization: Learning to Generalize to Unseen Domains}
\label{gen}
\input{PaperII/11_Generalization}

\section{Conclusion}
\input{PaperII/13_conclusions}

\section*{Acknowledgment of AI Assistance in Manuscript Preparation}
During the preparation of this work, the authors used ChatGPT to assist with refining and correcting the text. After using this tool, the authors carefully reviewed and edited the content as needed and take full responsibility for the content of this publication.
\bibliographystyle{unsrt}  

\bibliography{cleanedup}
\end{document}

%% file: PaperII/00_abstract.tex
Prognostics and Health Management (PHM) ensures the reliability, safety, and efficiency of complex engineered systems by enabling fault detection, anticipating equipment failures, and optimizing maintenance activities throughout an asset’s lifecycle. However, real-world PHM presents persistent challenges: sensor data is often noisy or incomplete, available labels are limited, and degradation behaviors and system interdependencies can be highly complex and nonlinear. Physics-informed machine learning has emerged as a promising approach to address these limitations by embedding physical knowledge into data-driven models. This review examines how incorporating learning and observational biases through physics-informed modeling and data strategies can guide models toward physically consistent and reliable predictions. Learning biases embed physical constraints into model training through physics-informed loss functions and governing equations, or by incorporating properties like monotonicity. Observational biases influence data selection and synthesis to ensure models capture realistic system behavior through virtual sensing for estimating unmeasured states, physics-based simulation for data augmentation, and multi-sensor fusion strategies. 
The review then examines how these approaches enable the transition from passive prediction to active decision-making through reinforcement learning, which allows agents to learn maintenance policies that respect physical constraints while optimizing operational objectives. 
This closes the loop between model-based predictions, simulation, and actual system operation, empowering adaptive decision-making. 
Finally, the review addresses the critical challenge of scaling PHM solutions from individual assets to fleet-wide deployment. 
Fast adaptation methods including meta-learning and few-shot learning are reviewed alongside domain generalization techniques that ensure robust performance across diverse assets and fleets. These advances are essential for moving PHM from isolated applications to broad, impactful deployment across real-world contexts. For each approach, we discuss its strengths, limitations, and relevance for different PHM tasks.

%% file: PaperII/01_Introduction.tex
Prognostics and Health Management (PHM) systems aim to ensure the long-term reliability, availability, and safety of complex industrial and infrastructure assets by predicting system degradation and supporting optimal maintenance decisions \cite{fink2020potential}. Increasingly,  PHM leverages machine learning (ML) techniques  for fault detection \cite{chao_hybrid_2019}, diagnostics \cite{wang2019domain}, and Remaining Useful Life (RUL) prediction~\cite{chao2022fusing}. However, practical  deployment of PHM   continues to face  persistent challenges: industrial data is often noisy or incomplete, labels are limited, and samples of faulty or degraded conditions may be partly or entirely missing. In addition, underlying degradation processes are frequently complex and nonlinear.  These obstacles  make it difficult for purely data-driven models to generalize beyond their training conditions  or produce physically consistent predictions. Overcoming these challenges requires approaches that combine the strengths of data-driven learning with insights grounded in the physics of the underlying systems.

\textbf{Physics-informed machine learning (PIML)} \cite{karniadakis_physics-informed_2021} has emerged as a powerful strategy to address these challenges by embedding physical knowledge into ML model structures and training. By integrating physical laws directly into the structure or training of machine learning algorithms, PIML approaches ensure that model predictions inherently respect known physical constraints, even when observations are limited. 
Within this context, three primary types of bias have been identified as essential for steering models toward physically consistent and reliable solutions: 

\begin{enumerate}
    \item \textbf{Inductive biases} are prior assumptions explicitly embedded into model architectures 
    \item \textbf{Observational biases} influence the model through  constraints  imposed on the selection,  representation and sampling of input  data, ensuring the data reflects relevant physical realities.
    \item \textbf{Learning biases} shape the training process itself, arising from the choice of algorithms, optimization objectives,  and regularization strategies. 
\end{enumerate}

This paper is the second part of a comprehensive review of physics-informed machine learning for PHM:
\begin{itemize}
    \item Part I explored  inductive biases, architectural design strategies  that embed  physical knowledge directly into model structures.
    \item Part II (this article) focuses on learning and observational biases, which influence the training process and data representation, as well as on scaling approaches that support robust deployment across diverse operational contexts.
\end{itemize}

\begin{figure}[t!]  
    \centering
    \includegraphics[width=0.99\textwidth]{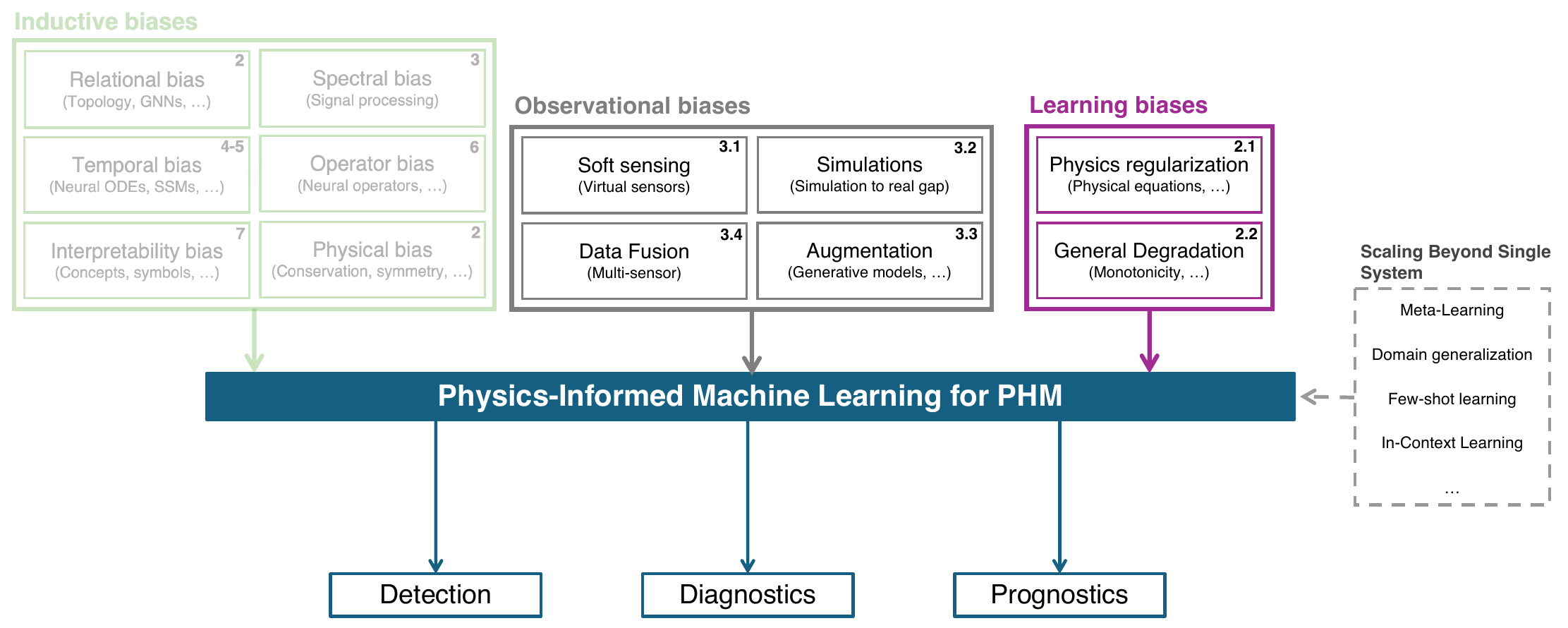} 
    \vspace{0.5cm}
    \caption{Physics-Informed Machine learning framework for PHM. Part I addressed inductive biases that embed structural assumptions into model architectures. Part II (this article) examines observational biases and learning biases that shape data representation and training dynamics. The framework extends to scaling approaches that enable fast adaptation and deployment across diverse systems.}
    \label{fig:big_figure}
    \vspace{0.5cm}
\end{figure}

Figure~\ref{fig:big_figure} illustrates how learning and observational biases complete the physics-informed framework for PHM.  This paper presents a comprehensive perspective  that begins with (1)  the role of  learning and observational biases in guiding models to learn physics-consistent system behavior from data, to (2) demonstrates  how these physics-informed predictions enable intelligent maintenance decisions through reinforcement learning, and (3) explores strategies for scaling  these approaches  to manage and optimize entire fleets of assets.

Across both parts of this review, we adopt a consistent structure: we first move from physics to machine learning by embedding physical laws into data-driven models, and then close the loop by returning from machine learning to physics-through the derivation of symbolic physical equations in Part I, and through interaction with the physical world via feedback and control in Part II.

\textbf{Learning biases} (see Section \ref{learning_bias}) are introduced through the careful selection of loss functions, constraints, and optimization algorithms, guiding machine learning models toward solutions that align with underlying physical laws. Unlike the architectural constraints of inductive biases examined in Part I, learning biases are explicitly imposed during model training to ensure that learned representations and functions remain consistent with known system behaviors.
In PHM applications, learning biases play a critical role in overcoming the fundamental challenge of data scarcity. Industrial datasets are often limited in size, coverage, or diversity, making it difficult for machine learning models to learn robust, generalizable patterns. By embedding physical laws and constraints directly into the learning process, learning biases enable models to compensate for missing or sparse data, effectively guiding them toward solutions that remain faithful to the underlying physics even when observations are incomplete. Without such guidance, purely data-driven optimization can lead models to overfit limited training data, resulting in predictions that violate basic physical principles~\cite{vafa2025what} or producing unrealistic degradation trajectories.

\begin{figure}[t!]  
    \centering
    \includegraphics[width=1.0\linewidth]{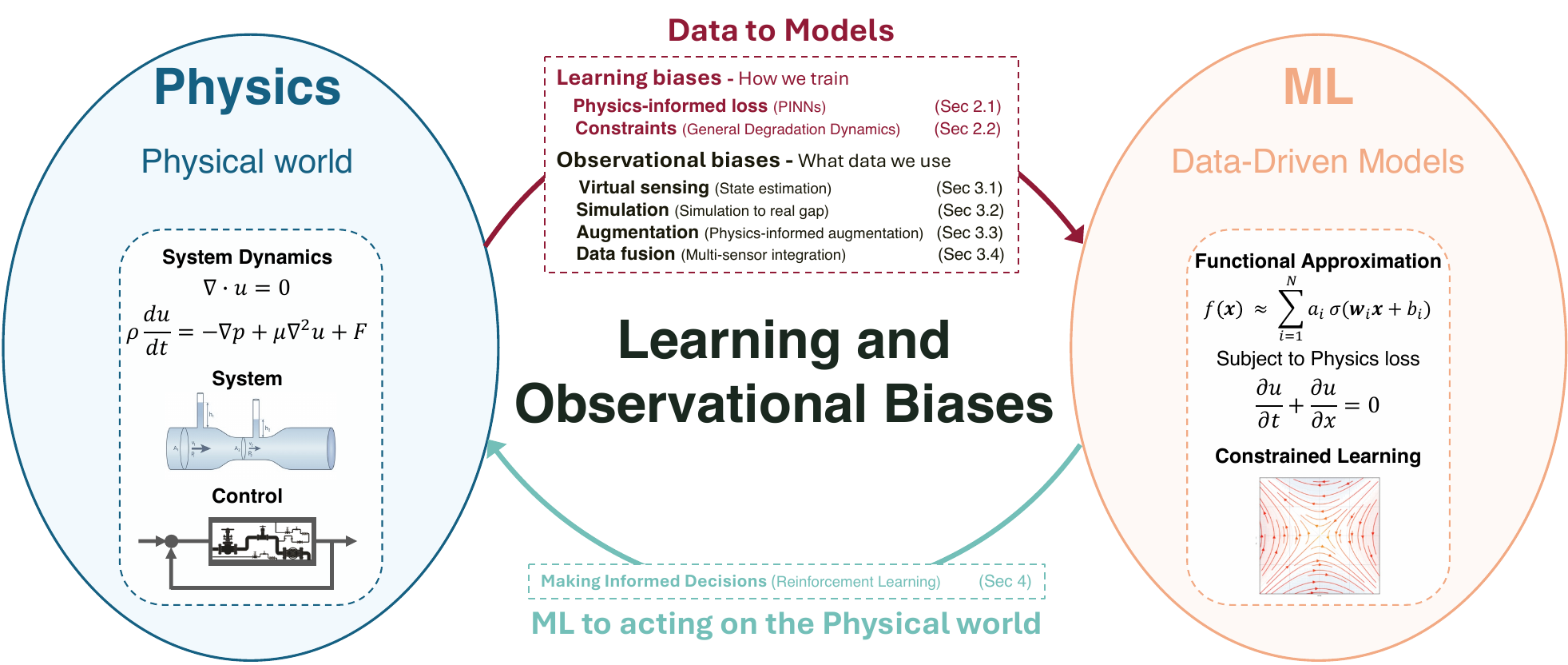}
    \caption{Overview of how learning and observational biases bridging Physics and Machine Learning within PHM framework. Observational biases (Section \ref{observational_bias}) ensure data reflects physical system realities, improving model generalization and robustness. Learning biases (Section \ref{learning_bias}) guide model optimization to produce physically consistent prediction. Reinforcement learning (RL)  (Section \ref{RL}) leverages these biases to transition predictive insights into actionable maintenance decisions, closing the loop between modeling and operational effectiveness in PHM.}
    \label{fig:cycles}
\end{figure}

In this review, we examine two primary mechanisms for introducing learning biases in PHM. Physics-Informed Neural Networks (PINNs) (see Section \ref{PINNS}), incorporate governing differential equations directly into the loss function, ensuring that model solutions satisfy conservation laws and system dynamics. As a complementary approach, General Degradation Dynamics Informed Prognostics Methods (see Section \ref{sec:soft_phys}) offer a more flexible alternative to physics-based or domain-specific approaches by embedding general physical properties, such as monotonicity and irreversibility,that characterize degradation processes across diverse systems. 

\textbf{Observational biases} (see Section \ref{observational_bias}) capture the patterns and distributions inherent  in the training data. 
These biases can be leveraged  not only by  curating datasets that both adhere to physical principles and comprehensively represent the system’s operational states, but also by enriching the data through virtual observations and simulated samples from conditions that were not originally observed. 
Conceptually, observational biases offer the most straightforward way  to embed physical knowledge into models --exposing them to physically meaningful data, thereby improving robustness.  However, PHM applications present unique challenges for data collection: sensors may be sparse or expensive, or impossible to install in certain locations; some  failure modes are rare; and specific  system states may be impossible to observe directly. These constraints often result in datasets with significant gaps and biases, which can mislead purely data-driven models. Physics-informed observational biases address these limitations through four key strategies.

Virtual sensing (see Section \ref{SS}) leverages known physical relationships to estimate unmeasured variables or reconstruct missing data using physics-based models, effectively expanding the observable state  space of the system. Physics-based simulations (see Section \ref{sim2real}) generate synthetic training data using first-principles models or numerical solvers, enabling the creation of representative samples for rare events  and a wider range of  operating conditions. Physics-informed data augmentation (see Section \ref{sec:generative}) increases the dataset diversity  by generating new data while strictly  preserving physical constraints. Finally, multi-sensor data fusion (see Section \ref{sec:data_fusion}) integrates  complementary sensor modalities to construct richer and  more complete representations of system behavior. Through these combined methods, models are exposed to data that more closely mirrors the true physical mechanisms of degradation. Table~\ref{tab:bias_advantages_limitations_part2} summarizes the key advantages and limitations of the learning and observational bias approaches discussed in this part of the review.

\begin{table*}[ht!]
\centering
\caption{Key advantages and limitations of learning and observational biases in PHM (Part II)}
\label{tab:bias_advantages_limitations_part2}
\begin{tabular}{p{0.5cm}p{2cm}p{6cm}p{6cm}}
\toprule
\textbf{Bias} & \textbf{Approach} & \textbf{Key Advantages} & \textbf{Main Limitations} \\
\midrule
\multirow{8}{0.5cm}{\rotatebox{90}{\textbf{Learning}}} &
\textbf{Physics-Informed NN (PINNs)} &
Improves generalizability via hard physics constraints; enforces physical consistency; reduces labeled data needs; extrapolation beyond observed regimes. &
High computational cost; hyperparameter sensitivity; scalability issues in high-dimensional PDEs; struggles with discontinuities. \\
& \textbf{General Degradation} &
Flexible embedding of general physical properties (monotonicity, trendability); applicable across diverse systems; less restrictive than PDEs. &
May oversimplify complex physics; requires domain knowledge for constraint selection; potential for over-regularization. \\
\addlinespace[0.7em]
\multirow{14}{0.5cm}{\rotatebox{90}{\textbf{Observational}}} &
\textbf{Virtual Sensing} &
Cost-effective alternative to physical sensors; estimates unmeasurable variables; provides redundancy for unreliable sensors; deployable in real time. &
Dependent on model fidelity and data quality; domain drift and calibration needs; limited transfer across operating regimes. \\
& \textbf{Sim-to-Real} &
Reduces reliance on labeled field data; covers rare/unsafe failure scenarios; enables safe and controllable data generation. &
Simulator fidelity limits transfer; domain gap/negative transfer risks; complex validation. \\
& \textbf{Generative Modeling} &
Generates rare-failure samples and augmentations; improves robustness and class balance; models data manifolds; adaptable across assets. &
Computationally intensive; synthetic–real domain gap; propagates dataset biases; training instabilities (e.g., mode collapse). \\
& \textbf{Data Fusion} &
Combines complementary modalities for richer health representation; improves accuracy and robustness; resilience to single-sensor failures. &
Heterogeneous data alignment/synchronization is challenging; sensitivity to missing modalities; calibration and scaling issues. \\
\bottomrule
\end{tabular}
\end{table*}

The learning and observational biases discussed above enable models to accurately capture system dynamics and degradation patterns in a way that is consistent with physical laws. Yet, effective PHM ultimately requires translating these \textit{passive predictions} into \textit{active decision-making} that interacts with and impacts the physical system itself.  This  transition  is achieved  through \textbf{reinforcement learning} (see Section \ref{RL}), which closes the loop by allowing agents to move from model-based predictions back to the physical world. By building on physics-informed representations, RL agents can learn maintenance and operational policies that not only balance operational constraints, costs, and  reliability, but also ensure that decisions are informed by the true physical behavior and degradation processes of the system.  

Finally, moving from individual assets to fleet-wide deployment introduces significant  \textbf{scaling challenges} (Section~\ref{scaling}) . To address these, we highlight fast adaptation approaches such as meta-learning and few-shot learning  (Section~\ref{sec:few_shot}), which facilitate rapid adaptation to new systems with limited data.  In addition, domain generalization methods (Section~\ref{gen})  are reviewed for their role in ensuring  models maintain  robust performance  across diverse assets and  operational contexts. These methods address the practical requirements of large-scale PHM, supporting reliable performance across multiple assets and varied operating environments.

Figure~\ref{fig:cycles} illustrates  our framework, which moves from physics to machine learning and then back to the physical system. Learning and observational biases  form the basis for  physics-informed predictions, while reinforcement learning closes the loop by translating these predictions into actions that directly affect  system behavior. The following sections  provide a detailed review of observational and learning biases, including their foundational principles, practical applications  in PHM, current limitations, and future  research directions. Integrating these systematic biases supports the development of  robust, reliable, and scalable PHM solutions, ultimately enhancing asset reliability and enabling  optimal operational decisions.

%% file: PaperII/02_MODEL_PINN.tex
\subsection*{Introduction to PINNs}
\label{sec:PINN}

\begin{table}[h!]
\centering
\belowrulesep=0pt
\aboverulesep=0pt
\renewcommand\arraystretch{2}
\begin{tabular}{@{}ccp{8cm}@{}}
\toprule
\multirow{2}{*}[-1ex]{Applications in PHM} &  Type of tasks         & Fault detection; RUL prediction. \\ \cmidrule(l){2-3} 
                                     & Addressed challenges  & Sparse data; Physical inconsistency; Poor generalization ability. \\ \midrule
\multirow{4}{*}{Requirements}        & Prior knowledge        & Governing differential equations.  \\ \cmidrule(l){2-3} 
                                     &  Data                  & Labeled data.             \\ \cmidrule(l){2-3} 
                                     & Type of bias           & Learning bias integrated through the loss function. \\ \cmidrule(l){2-3} 
                                     & Assumptions            & Smoothness.    \\ \midrule
\multirow{2}{*}[-4ex]{Summary}             & Advantages             & Improved generalizability; Incorporation of physical laws; Reduced dependency on large datasets. \\ \cmidrule(l){2-3} 
                                     & Disadvantages          &  Computational cost; Hyperparameter sensitivity; Scalability to high-dimensional problems;
                                     \\ \bottomrule
\end{tabular}
\vspace{0.5cm}
\caption{Checklist of \textit{Physics-Informed Neural Networks (PINNs)}}
\label{checklist:PINN}
\end{table}

Traditional neural networks (NNs) have become widely used in PHM applications due to their ability to capture complex patterns from data. However, high-quality labeled datasets are often scarce, as collecting real-world measurements or running detailed numerical simulations for complex systems is both costly and time-consuming. In these ``low-data'' settings, purely data-driven models face key challenges: First, NNs trained on limited data may struggle to generalize or converge at all, leading to unreliable predictions under new or varying conditions.

Second, standard  NNs can produce physically implausible outputs that violate essential  system constraints, such as symmetries and conservation laws, undermining  their reliability in PHM applications.

\begin{figure}[h]  
    \centering  
    \includegraphics[width=0.8\textwidth]{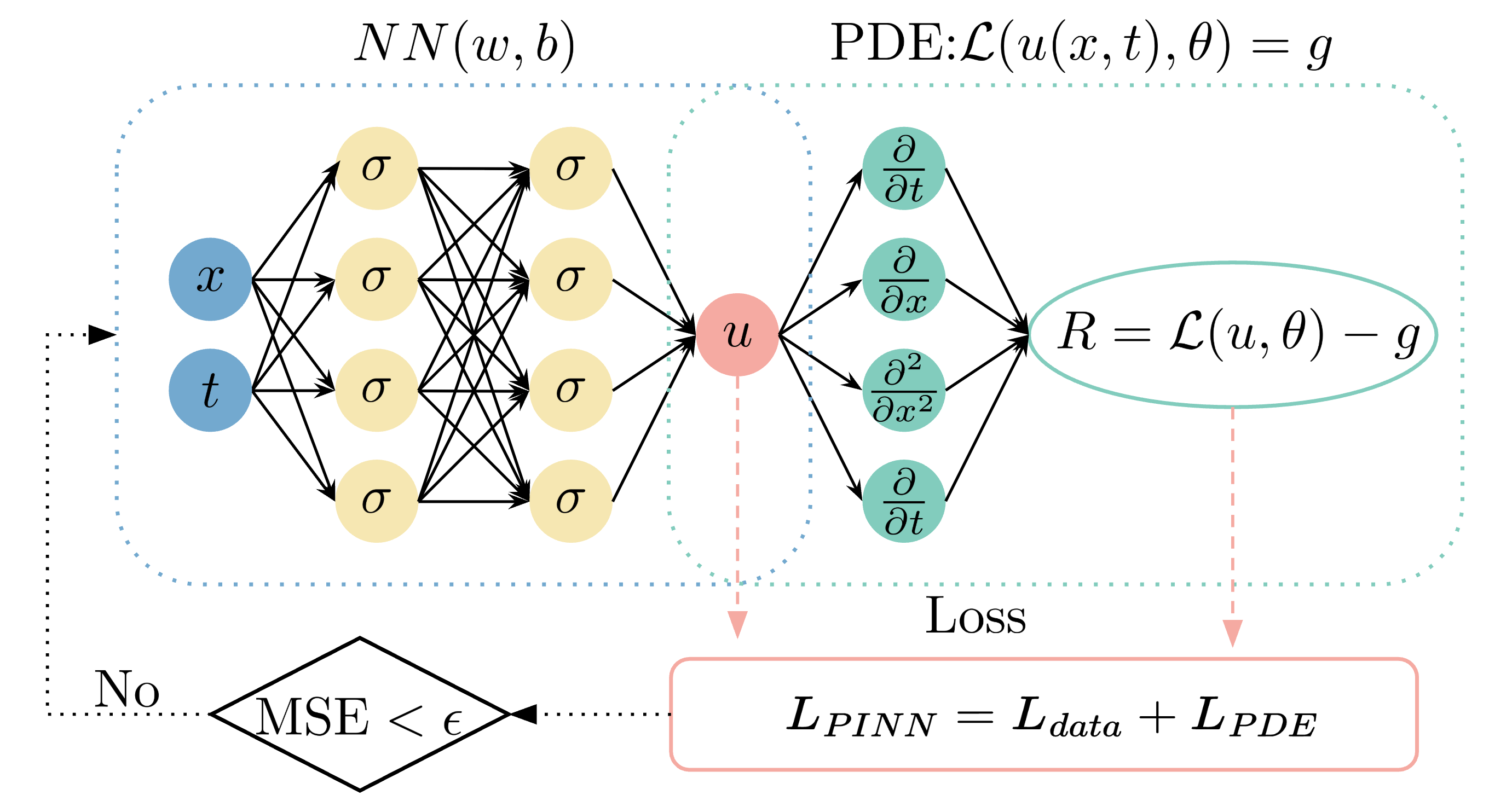}  
    \caption{Overview of a physics-informed neural network. PINNs are trained to directly approximate solutions to PDEs by encoding them as residues in physics-informed loss terms.}
    \label{fig:PINN}
\end{figure}

PINNs, first introduced by Raissi et al.~\cite{RAISSI2019PINNs}, address these challenges by embedding physical priors into the training process.  These priors, usually given in the form of governing partial differential equations (PDEs), are encoded as regularizing terms in a model's loss function. They then act in conjunction with inductive biases to produce predictions that are consistent with established physical principles, rather than relying exclusively on data-driven correlations. This makes PINNs a prototypical example of {learning bias},  

enhancing generalization in low-data regimes.

PINNs were first conceived as coordinate-based Multilayer Perceptrons (MLPs), used as universal approximators for solutions to PDEs. The key idea is to use these networks to directly model solutions, supplementing scarce data with domain knowledge by embedding differential equations into their training loss as residues.

For example, given a general  differential operator $\mathcal{L}$ and a source term $g$,  the physics-based constraint $\mathcal{L}(u) = g$ can be encoded as a \textit{physics-informed} loss:
\begin{equation}
\begin{aligned}
        &\mathcal{L}(u) = g \\
        \implies &L_{PDE} = ||\mathcal{L}(u) - g ||_2^2 = 0.
\end{aligned}
\end{equation}
which is minimized when the network’s prediction $u$ satisfies the underlying physical law.

Modern machine learning frameworks enable efficient computation of derivatives with automatic differentiation,   enabling the evaluation of such loss terms at randomly sampled collocation points throughout the domain. As illustrated in Figure~\ref{fig:PINN}, 
PINNs combine the physics-based loss with data-driven terms (e.g., initial / boundary conditions or measurement data):   
\begin{equation}\label{Eq:PINN_loss}
     {L}_{\text{PINN}} = w_{\text{data}} {L}_{\text{data}} 
    + w_{\text{PDE}}  {L}_{\text{PDE}} 
\end{equation}
where:
\begin{itemize}
    \item \( {L}_{\text{data}}\) represents the \textbf{data-driven loss}, which measures the discrepancy between the model’s predictions and actual observations, as well as boundary and initial conditions. 
    \item \( {L}_{\text{PDE}}\) enforces the \textbf{governing PDE constraints}, ensuring that the neural network satisfies the underlying physical equations at randomly sampled collocation points.
    \item \(w_{\text{data}}, w_{\text{PDE}}\) are weighting factors that balance the contributions of each loss component.
\end{itemize}

PINNs have since been extended in a number of ways, including schemes for domain decomposition \cite{Jagtap2020xPINNs} and Bayesian inference \cite{YANG2021BPINNs}. The concept has also been considered in a broader sense, leading to applications beyond conventional MLPs. Physics-informed constraints have been integrated as regularizers in a large number of modern architectures, including transformers \cite{ramirez2025residual}, and Neural ODEs \cite{sholokhov2023physics}. In this more general context, physics-informed terms may take the form of general operators, conservation laws, and symmetries, both continuous and discrete. These terms may encode a variety of physical priors, such as Hamiltonians \cite{mattheakis2022hamiltonian} and continuous symmetries \cite{zhang2023enforcing}. 

Integrating such {physical insight} directly  into  the training process reduces reliance on large labeled datasets, improves   {generalization, and  ensures that model predictions remain  consistent with known physical laws}.

\subsection*{Application of PINNs in PHM}
PINNs and their variants have seen a surge in interest in recent years, leading to a large body of work on PHM, including applications in detection, diagnostics and prognostics, where they contribute to improving the reliability of data-driven frameworks. While diverse, these applications are often based on combining observational data with a wide range of physics-based priors, such as conservation of energy. The concept of physics-informed (as well as "physics-inspired" and "physics-guided") networks has broadened in scope, leading to a diverse range of additional techniques for embedding physical insight into PHM model training, including physics-informed data augmentation~\cite{de2024remaining}. 
In \cite{sun2025damage}, a physics-informed term enforcing energy conservation is used to enable robust damage detection in wind fan turbines, using this regularization to refine unreliable measurements from acoustic sensors. In \cite{shukla2020physics}, PINNs are used to perform non-destructive testing by using the surface acoustic wave equation alongside ultrasound data to detect and describe surface cracks. In \cite{ramirez2025residual}, PINNs are used to track thermal insulation aging in power-plant transformers, using a PDE-based heat propagation model to extract detailed spatial thermal profiles from localized measurements. 

PINNs have also been used for multiple prognostics tasks. For instance, Zheng et al.~\cite{ZHENG2022PINN_RUL_crack} applied PINNs to predict crack growth in quasi-brittle materials in an unsupervised context. In this case, the loss function was formulated in terms of differential equations modelling damage increase and the minimization of elastic potential energy, leading to physically plausible crack configurations.  Similarly, Wen et al.~\cite{Wen2023PINN_RUL_Battery} developed a model fusion approach using PINNs for RUL predictions of Lithium-ion batteries. Here, the physics-informed loss encodes coarse low-dimensional approximations for degradation dynamics, while a data-driven model is used to model spatial dependences across the power cell. Additionally, Wang et al.~\cite{Wang2024PINN_RUL_Battery} introduced a PINN model for state-of-health estimation, whereby both the physics-informed equation residue as well as its solution are learned with neural networks, leading to a flexible framework for learning the unknown dynamics of battery degradation over time. 
A summary of the application contexts for PINNs is provided in Table \ref{checklist:PINN}.

\subsection*{Advantages and Limitations of PINNs}
These applications highlight the unique advantages of 

PINNs in PHM. By blending physics-based constraints with data-driven learning, PINNs enable more reliable and interpretable predictions, 

particularly in scenarios where high-quality labeled  data  are limited, costly, or noisy. 

Their ability to enforce physical consistency during training means that PINNs naturally denoise and correct sensor measurements, leading to  robust predictions that remain consistent with the underlying system dynamics. 
Unlike purely data-driven models, which often require large datasets to capture degradation patterns, PINNs leverage governing equations to guide learning. This allows them to predict fault onset, degradation progression, or remaining useful life accuratly even from limited or incomplete data.
Nevertheless, while they offer significant advantages in PHM, they also have several limitations that need to be addressed for broader adoption.

A significant limitation of PINNs is their sensitivity to hyperparameters and the challenge of loss balancing \cite{bischof2025multi}. Because  PINNs combine data-driven and physics-based losses, the choice of weighting factors is critical: poorly tuned weights factors can cause instability during training or lead to suboptimal solutions.  If the physics loss is weighted too heavily, the network may ignore available data; conversely, if the data loss dominates, the model risks losing physical consistency and behaving like a purely data-driven network. 
Another challenge is the requirement for explicit, accurate governing equations. In many practical  PHM applications, such as material fatigue or degradation prediction,  these equations may be incomplete, approximate, or entirely unknown, limiting the effectiveness of  PINNs. Furthermore, relying solely on physics-informed regularization is often insufficient to enforce physically plausible solutions.  In low-data regimes, PINNs  may fail to converge or produce trivial, non-physical outputs. This is particularly prevalent for systems dealing with high-frequency behavior or complex geometries, restricting PINNs to relatively simple equations. This often requires the introduction of additional inductive biases, such as specialized neural  architectures \cite{bizzi2025neural} or tailored regularization strategies.

Despite these limitations, ongoing research in adaptive loss balancing, hybrid modeling approaches, and more computationally efficient  training algorithms continues to broaden  the applicability and scalability of PINNs in PHM.

\subsection*{Unexplored or Emerging Applications of PINNs in PHM}

Although Physics-Informed Neural Networks have been widely used across multiple sub-domains of PHM, they still hold considerable untapped potential, particularly in domains where classical degradation models exist but are computationally expensive, incomplete, or difficult to integrate with real-world data. 

So far, in part due to the limitations of PINNs, the physics-informed losses used often refer to relatively basic, phenomenological descriptions of the system. For this reason, they still cannot be considered replacements for traditional numerical simulators when complex geometries or dynamics are at play. Likewise, the application of physics-informed terms to the training of emerging specialized architectures remains a promising field in general.

In terms of applications, the analysis of tribological wear prediction in gears and bearings shows promise. Traditional wear models, such as Archard’s law, provide general insights but fail to capture complex tribological interactions like nonlinear friction and lubrication effects \cite{varenberg2022adjusting}. PINNs have been used in an unsupervised manner in this context \cite{marian2023physics} and could be used further to integrate physical laws with sensor data to improve the real-time estimation of wear rates. 
In corrosion and material degradation modeling, PINNs could enhance the understanding of time-dependent electrochemical reactions affecting structures like pipelines, aircraft, and marine vessels. Modeling corrosion-fatigue has proven to be hard, which has led to the development of hybrid data-driven frameworks \cite{Dourado2019_PImodel}. PINNs could be used in conjunction with real-world data to account for environmental variability (e.g., humidity,
salinity) and provide more robust and site-specific degradation forecasts.

Finally, PINNs could be used to integrate a wider range of physical priors with non-destructive testing (NDT). NDT often involves the interaction of multiple complex and interdependent physical phenomena, which can be described by elaborate PDE systems. PINNs could improve the detection of hidden subsurface defects using ultrasonic, eddy current, or infrared thermography data, along with the mathematical descriptions thereof, leading to more reliable condition assessments and defect identification in aerospace, civil, and industrial structures.

\subsection*{Promising Research Direction for PHM: Physics-Informed Computer Vision}
\input{PaperII/03_PICV_Promising_Direction}
\label{sec:pinn_visual}

%% file: PaperII/03_PICV_Promising_Direction.tex
A particularly promising direction for extending PINNs in PHM is Physics-Informed Computer Vision (PICV). Visual data is increasingly central to PHM applications, particularly for inspection tasks such as defect detection \cite{malekzadeh_aircraft_2020}, structural monitoring \cite{cha_deep_2017,li2024road}, and visual condition assessment \cite{mishra_artificial_2024}. With the growing use of autonomous platforms (e.g., drones, underwater vehicles, robotic arms), computer vision (CV) enables scalable, non-intrusive monitoring across diverse assets and environments.

Despite this potential, current CV models in PHM are often limited by their reliance on large labeled datasets and poor generalization across operating conditions. Many models are trained for specific scenarios and may underperform under variable viewpoints, lighting, or noise\textemdash frequent challenges in real-world deployments. PICV aims to address these limitations by embedding physical constraints into the learning process, improving robustness, and interpretability in data-scarce or safety-critical settings. This integration of physics with vision offers a natural extension of PINNs to visual modalities, opening new opportunities for physics-consistent PHM systems.

PICV integrates domain knowledge into data-driven CV models by embedding physical priors through three main types of biases \cite{karniadakis_physics-informed_2021,banerjee2023}:
(1) Observational biases: capturing physical quantities directly through enriched sensing modalities (e.g., depth or thermal imaging \cite{huo_glass_2023,elsayed_savi_2022});
(2) Learning biases: regularizing training using physical equations (e.g., heat transfer, Navier-Stokes) or geometric constraints (e.g., epipolar geometry, photometric consistency) to improve data efficiency and enforce physically plausible outputs \cite{gao_super-resolution_2021,cai_flow_2021};
(3) Inductive biases: embedding symmetries and invariances (e.g., rotational, scale) into architectures using group convolutions or steerable CNNs \cite{cohen_group_2016,bronstein_geometric_2021}. These strategies typically enhance model robustness, learning efficiency, and improve physical consistency, particularly beneficial for PHM tasks where deployments face highly variable conditions. In label-scarce scenarios, common in PHM, these physics-based objectives can also be applied in a self-supervised manner, allowing models to learn from abundant unlabeled visual data while maintaining physical consistency. For an extensive survey on PICV, we guide the reader to \cite{banerjee2023}.

While CV is widely used in PHM across domains\textemdash such as railway inspection \cite{gibert_deep_2017}, aerospace surface analysis \cite{malekzadeh_aircraft_2020}, and civil infrastructure monitoring \cite{cha_deep_2017,li2024road}\textemdash the explicit use of PICV for PHM remains limited. A few hybrid approaches exist, such as combining grease condition analysis via vision with physics-based models for bearing degradation in wind turbines \cite{yucesan_hybrid_2021,yucesan_hybrid_2022}, but these are still rare.

PICV opens promising research directions in PHM. Potential applications include dynamic process monitoring, where vision-based fluid flow or thermal field modeling could enable real-time anomaly detection (e.g., cavitation, heat build-up) \cite{gao_super-resolution_2021,zhao_physics-informed_2023}. Additionally, equivariant architectures (e.g., to rotation and scale) can improve inspection robustness for structures and components \cite{mishra_artificial_2024}. Finally, combining vision with sensor data (e.g., vibration, pressure)\textemdash a strategy known as multimodal fusion (see Section~\ref{sec:data_fusion})\textemdash can lead to more comprehensive health assessment systems \cite{9720937,Wang_2023_CVPR}.

Despite these opportunities, PICV faces practical challenges, many of them similar to those faced by PINNs. Designing physics-consistent architectures requires domain knowledge in physics (e.g., PDEs, system symmetries), which may be complex to formalize. Some PICV approaches require explicit governing equations, which may be hard to derive or too simplified for real-world PHM. Vision-specific issues also arise: physical constraints must be enforced in image space, where camera projection effects, occlusions, and lighting variations complicate modeling. Finally, PICV methods often require synchronizing visual and physical data across modalities, adding complexity to training and deployment.

%% file: PaperII/04_SOFT_PHYS.tex
\subsection*{Introduction to Degradation Dynamics-Informed Prognostics Methods}

Industrial and infrastructure assets typically undergo progressive degradation over time due to wear, fatigue, thermal stress, and other environmental and operational factors. Generally, such degradation can be modeled using partial differential equations or other first-principles approaches. For example, in the case of wear, PDEs may capture erosion depth as a function of applied loads and microstructural properties \cite{cubillo_review_2016}. Similarly, battery degradation can be modeled through electrochemical reaction kinetics \cite{reniers2019review}, and turbine blade erosion can be described by stress-strain relationships \cite{yu2017new}. However, for complex systems comprising multiple interacting components and difficult-to-measure parameters (such as friction coefficients, stiffness, damping, contact forces, and material inhomogeneities) PDE-based degradation models become impractical \cite{peng2022mechanistic}. This is due to the high dimensionality arising from the need to solve coupled PDEs across different components, the strong nonlinear interactions between these equations, and the limited accessibility of internal variables such as internal loads and stresses that are required for model calibration \cite{cubillo_review_2016}.

Instead of directly modeling degradation through first principles, many studies rely on \textit{health indicators}, quantities that aggregate the level of degradation based on sensor data \cite{liu2013data,xu2021machine}. These may be physically interpretable (e.g., crack length, battery capacity) or latent variables learned from multi-sensor signals. The learning process depends on the nature of available data: when direct measurements of degradation are available, health indicators can be learned in a supervised manner~\cite{roman2021machine}; otherwise, semi-supervised or unsupervised techniques are employed to uncover underlying degradation trends from unlabeled data \cite{wen2021generalized,frusque_semi-supervised_2024,liu2019remaining}.

\paragraph{Monotonicity}
Irreversible degradation processes—such as crack growth or battery capacity decline—can be modeled by enforcing a monotonic change over time in the health indicator.  This can be imposed using a penalty term that discourages non-monotonic transitions \cite{liu2013data,wen2021generalized}:
\begin{equation}
L_{\text{mono}} = \sum_{t=1}^{N-1} \max(0, H_t - H_{t+1})
\end{equation}
where $H_t$ is the health indicator at time $t$. This loss, designed for decreasing health indicators, penalizes instances where the indicator increases (i.e., when $H_t > H_{t+1}$) thereby discouraging non-monotonic behavior.
 
\paragraph{Trendability}
Degradation processes typically evolve consistently over time or with accumulated usage. For instance, bearing degradation should correlate with the number of revolutions, and battery degradation should align with charge-discharge cycles. To reflect this, health indicator can be enforced to correlate with the operational cycle index $C$ \cite{wen2021generalized,bajarunas2024health}:
\begin{equation}
L_{\text{trend}} = -\rho(C, H)
\end{equation}
where $\rho(C, H)$ measures the degree of correlation. 

\begin{figure}[ht]  
    \centering  
    \includegraphics[width=1.0\textwidth]{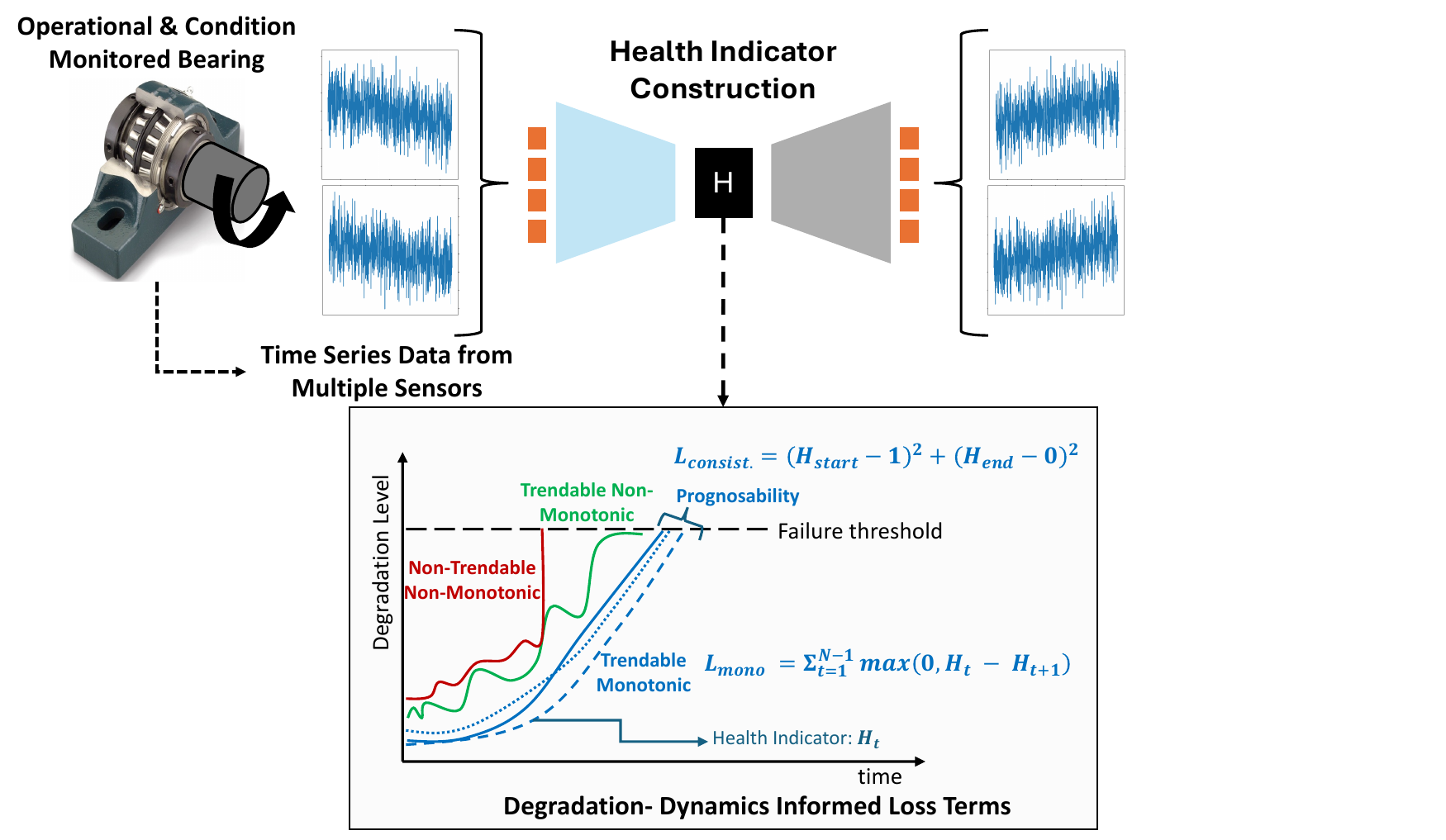}  
    \caption{Overview of degradation dynamics-informed prognostic methods. Time-series sensor data from monitored systems (e.g., bearings) are processed to construct health indicators that progressively satisfy key degradation properties: monotonicity (irreversible decline), trendability (correlation with operational cycles), and Prognosability (comparable endpoints across similar units). Auxiliary loss terms enforce these physics-motivated constraints during the learning process, guiding the evolution from non-physical to physically meaningful health indicators.} 
    \label{fig:soft_phys}
\end{figure}

\paragraph{Prognosability}
Systems undergoing the same degradation process should exhibit comparable health trajectories that enable consistent prognostic performance. 
When the health indicator is scaled between 0 (healthy) and 1 (failure), the health indicators should converge to similar values at key degradation stages, particularly at failure. 
To enforce this, a penalty can be applied to align the indicator values at the start and end of the degradation trajectory \cite{li2020shape,chen2021health}:
\begin{equation}
L_{\text{consistent}} = (H_{\text{start}} - 1)^2 + (H_{\text{end}} - 0)^2.
\end{equation}

\paragraph{Robustness}
Degradation patterns are typically robust to noise and fluctuations in sensor readings caused by environmental or operational variability. To reflect this, the learned health indicator can be constrained to remain smooth over time, suppressing high-frequency variations that do not correspond to actual degradation. This behavior can be encouraged using regularization or temporal smoothing techniques \cite{yan2020generic,chen2021health}:
\begin{equation}
L_{\text{robustness}} = \frac{1}{N} \sum_{t=1}^{N} \exp \left( -\left| \frac{H_t - \bar{H}}{\bar{H}} \right| \right)
\end{equation}
where $\bar{H}$ is the mean health indicator over time. This loss penalizes values that deviate from the mean of the health indicator over time.

For modeling a system's degradation process, one or more of these loss formulations can be selectively applied depending on the available data, domain knowledge, and desired properties of the health indicator.

\subsubsection*{Application of Degradation-Dynamics-Informed Methods in PHM}
Recent studies have incorporated degradation dynamics into the learning process by embedding monotonicity constraints in the health indicator formulation along with supervision on the underlying measured parameters. For example, Kim et al.~\cite{kim2022data} used simulated crack growth data based on Paris’ law to constrain a neural network to follow the irreversible progression of damage, resulting in strong agreement with known crack propagation trends. 
Building on similar physics-motivated constraints, Chen et al.~\cite{chen2022physics} developed a degradation-consistent recurrent neural network for bearing prognosis using temperature signals. By imposing a monotonic trend on predicted temperature values, their method improved the temporal consistency of degradation modeling and yielded more stable remaining useful life estimates.

Other studies have employed unsupervised or semi-supervised approaches to learn latent health indicators from multi-sensor data by imposing degradation-informed constraint, particularly in settings with no direct measurement of degradation. For example, Wang et al.~\cite{wang2020deep} proposed a deep learning–based sensor fusion method that integrates signals such as vibration and acoustic emissions, and applies monotonicity and range constraints to produce a consistent indicator of degradation. Their approach was demonstrated on simulated turbofan engine degradation data. Similarly, Chen et al.~\cite{chen2021health} developed a feature fusion framework for lithium-ion batteries and rolling element bearings, optimizing the combination of degradation-relevant features while enforcing monotonicity, trendability, and robustness. Following a similar idea, Bajarunas et al.~\cite{bajarunas2024health} introduced degradation-informed learning biases (such as trendability and negative gradient constraints)into an autoencoder architecture to infer latent degradation signals from turbofan and battery datasets. Wen et al.~\cite{wen2021generalized} constructed a signal-based indicator for aircraft turbine engines by optimizing for trajectory consistency and enhanced signal range across degradation cases. More recently, Qin et al.~\cite{qin_new_2023} proposed a supervised multi-head self-attention autoencoder for machinery prognostics, using power function–inspired constraints to improve similarity-based remaining useful life prediction for wind turbine gearboxes and aero engines. These studies demonstrate the utility of degradation-consistent learning objectives for inferring health indicators in complex systems where direct measurements are limited or unavailable.

\subsection*{Advantages and Disadvantages of Degradation-Dynamics-Informed Methods in PHM}
Degradation-dynamics-informed methods incorporate generic constraint into the learning objective to better capture the intrinsic progression of degradation. These inductive biases are particularly beneficial in high-dimensional, multi-sensor settings, where each sensor provides only partial and potentially noisy observations of the underlying degradation process that is typically not directly measured or known \cite{bajarunas2024health,yan2020generic,xu2021machine}. By embedding these constraints, such methods improve robustness to sensor noise, mitigate overfitting, and enhance the interpretability of the learned health indicator. 

Owing to the generic nature of these learning biases~\cite{yan2020generic}, degradation-dynamics-informed methods are broadly applicable across diverse domains—ranging from bearings~\cite{sun2025unsupervised} and batteries~\cite{khaleghi2019developing} to turbine engines~\cite{bajarunas2024health}—thus providing another key advantage for practical deployment in PHM.

However, these methods also present notable challenges. A key limitation lies in their reliance on the assumption of monotonic degradation, which does not universally hold. Systems subject to resets (e.g., maintenance events), regime changes, or operational variability may exhibit non-monotonic degradation trajectories~\cite{caballe2015,tajiani2024optimizing,liang2023reliability}. In particular, systems exposed to random external shocks or varying operational loads deviate from smooth, unidirectional degradation patterns, thereby violating the foundational assumptions of these methods. 
Moreover, the varying operating conditions can make it difficult to consistently impose these constraints across different operational regimes, potentially leading to inaccurate prognostic outcomes when the assumed degradation dynamics do not align with the actual system behavior.

\subsection*{Unexplored or Emerging Applications}

Degradation-dynamics-informed prognostic methods have shown strong performance in established applications such as bearings~\cite{sun2025unsupervised}, batteries~\cite{khaleghi2019developing}, and turbofan engines~\cite{bajarunas2024health}. 
However, their application to real industrial data collected under realistic operating conditions remains limited, with many studies relying on controlled experimental setups or simulated datasets (e.g., NASA's turbofan simulation data \cite{4711414}). Moreover, their use in multi-component systems and more complex variable industrial environments is underexplored. Emerging domains such as wind turbines, railway components (e.g., axle bearings and brake pads)~\cite{ma2024change}, and industrial robotic systems~\cite{li2025trend} operate under non-stationary conditions involving varying loads, control strategies, and maintenance events. These factors often lead to non-monotonic degradation patterns that violate assumptions underlying many current methods.

To address these challenges, future research should focus on combining \emph{regime-aware segmentation} with \emph{partial-monotonic constraints}, or explicitly incorporating control inputs into the learning architecture. By identifying operational segments where degradation evolves consistently (using tools such as change point detection, load profile analysis, or maintenance records) constraints like monotonicity or trendability can be selectively enforced. 
Prior work has highlighted the importance of jointly modeling gradual degradation and shock-induced damage, particularly in multi-component systems where failures may result from both cumulative wear and abrupt events \cite{caballe2015,dong2021reliability}.

Integrating such physics-informed and shock-aware mechanisms into deep learning frameworks—as auxiliary loss functions—can enhance the robustness, interpretability, and adaptability of learned health indicators. These advances are essential for deploying degradation-dynamics-informed methods in realistic industrial environments characterized by variable operating conditions, irregular maintenance, and noisy sensor data.

%% file: PaperII/05_METHODS_SOFTSENSING.tex
\begin{table}[ht!]
\centering
\belowrulesep=0pt
\aboverulesep=0pt
\renewcommand\arraystretch{2}
\begin{tabular}{@{}ccp{8cm}@{}}
\toprule
\multirow{2}{*}[-1ex]{Applications in PHM} & Type of tasks         & Fault detection and Prognostics. \\ \cmidrule(l){2-3} 
                                     & Addressed challenges  & Lack of direct measurements; Sensor costs; Sensor failure or degradation; Estimation of difficult-to-measure or unmeasurable variables; Data sparsity or missing values. \\ \midrule
\multirow{3}{*}{Requirements}        & Prior knowledge        & Optional for Data-Driven: System dynamics; Physics-based models.  \\ \cmidrule(l){2-3} 
                                     & Data                  & Labeled data; Historical sensor measurements.
                                      \\ \cmidrule(l){2-3} 
                                     & Type of bias                  & Observational bias.\\ \cmidrule(l){2-3} 
                                     \cmidrule(l){2-3} 
                                     & Assumptions            & System behavior can be approximated through mathematical or data-driven models.    \\ \midrule
\multirow{2}{*}[-4ex]{Summary}             & Advantages             & Cost-effective alternative to physical sensors; Can estimate unmeasurable variables; Backbones for unreliable physical sensors. \\ \cmidrule(l){2-3} 
                                     & Disadvantages          & Sensitivity to model inaccuracies in Model-Driven methods, Dependence on data quality; Potential interpretability issues in deep learning models. \\ \bottomrule
\end{tabular}
\vspace{0.5cm}
\caption{Checklist of \textit{Soft Sensing}}
\label{checklist:SoftSensing}
\end{table}

\subsection*{Introduction to Soft Sensing}

\begin{figure}
    \centering
    \includegraphics[width=0.9\linewidth]{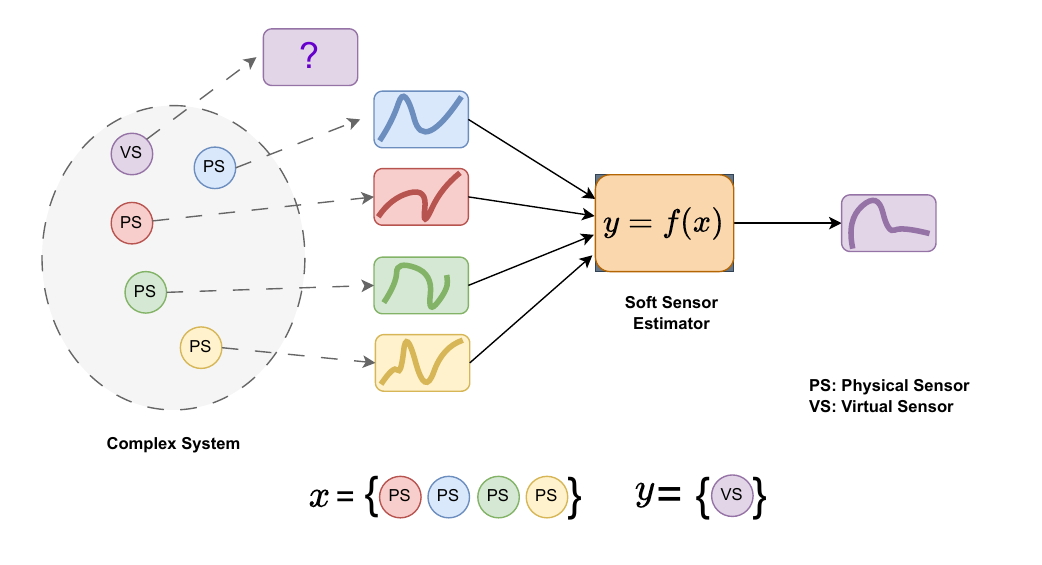}
   \vspace{-0.7cm}
    \caption{Illustration of a complex system with both physical sensors (measuring real-time parameters) and virtual (soft) sensors. The diagram highlights the mapping process from physical sensors to soft sensors using deep learning or other predictive modeling techniques, enabling enhanced monitoring, fault detection, and system optimization.}
    \label{fig:Softsensingmethods}
\end{figure}

In industrial settings, sensors are widely used to monitor processes, assess health condition, and ensure asset reliability. Ideally, placing sensors at every critical location would provide the most comprehensive process insight. However, practical and economic constraints often make this impossible. Some areas are inaccessible or exposed to harsh environments, where installing or maintaining sensors is not feasible.  Additionally,  retrofitting existing systems to accommodate new sensors can be prohibitively expensive and disruptive, often requiring production downtime, equipment modifications, or significant re-engineering. As a result, many important system variables cannot be measured directly by physical instruments alone. To address  these challenges, soft sensing (also known as virtual sensing) has emerged as a practical alternative. Soft sensing uses computational  and domain knowledge  to estimate key process variables  or product quality  based on available measurements \cite{perera2023role}, as illustrated  in Figure \ref{fig:Softsensingmethods}. This approach offers a cost-effective solution when  physical sensing is impractical or prohibitively expensive \cite{jiang2020review}. Selecting  the right soft sensor model is crucial  for ensuring accurate predictions and effective system monitoring. The two main  approaches for developing  soft sensors are white-box (model-driven) and black-box (data-driven) modeling, each with specific advantages and limitations \cite{kadlec2009data}.

\textbf{Model-Driven Soft Sensing Methods:}  Model-driven, or white-box, soft sensing approaches  are grounded in fundamental  principles of physics, chemistry, and engineering that govern system behavior \cite{torgashov2019use}. Common  techniques  include first-principle models, Kalman filters, and observers \cite{perera2023role}. 
\textbf{First-principle models} are particularly effective  when the governing equations of a system are well-understood and can be mathematically formulated, providing high interpretability through conservation laws \cite{bikmukhametov2020first}. Their main limitation, however, is the need for comprehensive system knowledge. 
\textbf{Kalman filters} provide a complementary approach when a complete mechanistic description of the system is not feasible (see Section 4 of Part 1). These estimation algorithms fuse noisy sensor measurements with simplified system models, which may be based on linearized versions of  nonlinear dynamics, mass or energy balances under steady-state assumptions, or empirical relationships derived from experimental data.
By capturing the essential system behavior without requiring a full mechanistic model, Kalman filters infer unmeasured states through sequential estimation (see Section \ref{sec:data_fusion}), continuously updating their predictions as new measurements arrive to enable real-time state tracking \cite{khodarahmi2023review, alexander2020challenges}.

\textbf{Data-Driven Soft Sensing Methods:} Data-driven approaches  construct soft sensing   models directly from empirical data, without relying on explicit physical models of the underlying  processes \cite{cao2025comprehensive}. They are particularly valuable in industrial contexts where where detailed process models are unavailable or or incomplete, but  large volumes of  operational data are routinely collected. 

Historically, linear and piecewise linear models have been the foundation of data-driven soft sensor modeling \cite{poggi2011high,sefati2019fbg}. Among these, Partial Least Squares (PLS) is a widely used method in industrial practice  due to its robustness to measurement noise and multicollinearity \cite{wang2015comparison}. However, the linear assumptions of PLS limit its capacity to represent complex, nonlinear system dynamics. To address  this, nonlinear approaches such as Support Vector Regression  have been explored  \cite{kaneko2014application, shokri2015improvement, liu2020soft}, along with probabilistic approaches like Gaussian Process Regression, which can capture nonlinear relationships while also providing uncertainty estimates \cite{liu2015auto,sheng2020smart}. Nonetheless, these machine learning methods typically require extensive feature engineering, which often falls short of fully capturing the complexity of underlying process behaviors.

Recently, deep learning-based techniques have gained traction for soft sensing, as they can automatically learn relevant features and representations from raw data. Unlike conventional methods, deep learning reduces reliance on expert-crafted features \cite{sun2021survey,gallareta2025advancements}. 
This shift has opened new possibilities for modeling highly complex and nonlinear industrial processes.

To capture spatial correlations in high-dimensional process data, convolutional neural networks (CNNs) have been applied to extract multiscale spatial features, with several studies demonstrating in taking advantage of the correlations between process variables \cite{wang2019dynamic,song2024soft}. To model temporal correlations, gated recurrent units (GRUs) have been adopted to capture sequential dependencies in process data, with bidirectional and hybrid designs improving predictive accuracy and robustness \cite{9865862,tian2022novel}. 

However, these conventional deep learning models generally assume fixed data structures and may not fully capture the irregular, dynamic, and relational nature of complex sensor networks.

To address these limitations, both transformers and graph neural networks (GNNs), including their extension Spatial-Temporal Graph Neural Networks (STGNNs) (see Section 2 of Part 1) have been explored. Transformers are well suited for capturing global, long-range dependencies through attention mechanisms~\cite{Zhang_2021}, whereas GNNs explicitly encode relational inductive biases by leveraging the topology of sensor networks. STGNNs further extend this capability by jointly modeling spatial structures and temporal dynamics~\cite{yang2025smart}.  

For example, Felice et al. \cite{felice2024graphbased} proposed a graph-based deep learning architecture for soft sensing in sparse multivariate sensor networks. It models both spatial and variable-wise dependencies using a nested graph structure: an inter-location graph captures learned similarities between sensor sites, while an intra-location graph models relationships among sensor variables at each location.  
Building on the idea of explicitly modeling structural dependencies, Zhao et al. \cite{zhao2024virtual} extend this line of work by introducing a GNN‑based soft sensing method that explicitly models sensor heterogeneity. Their heterogeneous temporal GNN treats different sensor types (e.g., temperature vs. vibration) as distinct node types, applies modality-specific temporal encoders, and incorporates operating condition context. 
Another important advancement in soft sensing is the incorporation of physical laws and domain knowledge directly into the learning process. For example, Niresi et al.\cite{faghih2024} propose a framework that augments the graph structure with auxiliary nodes and connections derived from governing equations, embedding process physics into the model architecture. Adding auxiliary nodes to the graph structure allows the GNN to map from a more enriched input space to the soft sensors, which improves estimation accuracy. Moreover, physics-guided graph learning combines a topology informed by known process relationships with learned correlations, and constrains the training objective through physics-based regularization terms that penalize violations of conservation constraints \cite{liu2024physics}, which leads to more accurate soft sensor estimation. Beyond graph-based methods, physically informed hierarchical learning \cite{aina2025physically} integrates reduced-order physical models into a multi-level neural architecture, using them as inductive biases to structure intermediate representations and guide parameter estimation when the underlying PDE dynamics are only partially known, further enhancing soft sensing performance.

\subsection*{Applications in PHM}

In PHM, soft sensing serves multiple critical functions that enhance system reliability and reduce operational costs. Its applications span diverse domains, including process industries, mechanical systems, structural health monitoring, and energy systems, each with unique measurement challenges and data characteristics.

\textbf{Estimation of Unmeasurable Variables:} Soft sensors enable the estimation of variables that cannot be directly measured due to harsh environments, physical inaccessibility, or prohibitive costs. In process industry, they are widely used to monitor complex product quality variables such as chemical component concentrations, polymer density, and melt flow rate, which are essential parameters  for precise control and optimization \cite{funatsu2016soft}. Similarly, in nuclear energy systems, virtual sensing predicts critical thermal-hydraulic characteristics under extreme conditions \cite{hossain2024virtual}. Beyond process environments, 
structural health monitoring is another key domain for soft sensing. In bridges, limited sensor deployments often constrain direct measurement of variables such as strain, displacement, or real-time load. The number and placement of sensors are often further restricted by costs, deployability, or accessibility, which motivates the use of virtual sensing approaches. Virtual sensors can infer these unmeasured responses by fusing heterogeneous sensing modalities with environmental and loading context. Zhao et al.\cite{zhao2024virtual} propose a heterogeneous temporal GNN that models distinct sensor types, integrates operating condition context, and predicts both bearing loads in rotating machinery and real-time loads in bridges, demonstrating the cross-domain applicability of advanced GNN-based virtual sensing.
Virtual sensing was also applied to electromechanical assets, with electric motors employing soft sensors to infer internal winding temperatures and insulation aging \cite{kral2013practical, kim2022stator}. Moreover, energy storage systems utilize virtual sensing in battery management to derive crucial state-of-charge  and state-of-health estimates from operational voltage, current, and temperature signals \cite{dini2024review}. These diverse applications highlight the adaptability of soft sensing across domains, enabling the indirect measurement of critical variables that would otherwise remain inaccessible.

\textbf{Backup Measurements for Sensor Failure Prevention and Detection:} Soft sensors serve also as backup measurements that maintain operational continuity when physical sensors fail, preventing costly shutdowns. In addition, they support the detection of faulty or degraded sensors.

For example, Darvishi et al. \cite{darvishi2020sensor} proposed a data-driven modular framework, in which the available sensor signals are partitioned into a reliable set and an unreliable set. 
Multilayer perceptrons are trained to model each signal in the unreliable set using only the reliable signals as input. By comparing the predicted (virtual) signals to the actual sensor outputs, the system performs residual analysis to detect faults.
In a different application, district heating networks rely on differential pressure to monitor pump performance, identify flow imbalances, and detect faults such as blockages or leaks. Accurate measurement of this parameter is essential for optimizing control strategies and maintaining energy efficiency, and data-driven soft sensors are employed to infer differential pressure values \cite{frafjord2024data}. Moreover, in HVAC systems, soft sensors support fault tolerance (i.e., maintaining reliable system outputs despite sensor faults) by detecting sensor anomalies through residual analysis and maintaining reliable outputs using soft sensors that are trained based on historical data \cite{nie2023sensor}. This ensures continued operation and control accuracy despite sensor failures, improving system resilience and reducing maintenance needs.

\textbf{Data Augmentation for Downstream Tasks:} Soft sensing can also expand the set of input variables by inferring additional unobserved quantities that can be used for more accurate and more interpretable fault detection, isolation and prognostics \cite{arias2019hybrid}.  For example, Arias et al. \cite{chao2022fusing} employ a calibrated physics-based performance model of a turbofan engine to estimate latent health-related parameters and virtual signals, which are then combined with sensor measurements as inputs to a deep learning model. This hybrid approach enhances prognostic performance while reducing dependence on large training datasets. Moreover, Niresi et al.\cite{faghih2024} augment the graph structure with auxiliary nodes representing pressure drop and temperature drop, demonstrating that this enhancement improves the estimation accuracy of the outputs which are pressure and temperature values at each point of district heating networks.

A summary of the application contexts for soft sensing is provided in Table \ref{checklist:SoftSensing}.

\subsection*{Advantages and Limitations of Soft Sensing}
 
Soft sensing has become vital in PHM, where it enables estimation of health indicators and process variables that are difficult or impossible to measure directly. Its main advantages include:
(1) \textbf{Enhanced observability:} estimation of variables that cannot be directly measured, including internal degradation states, operating loads that influence degradation, process conditions at additional spatial locations, and other critical unmeasured variables.
(2) \textbf{Cost and feasibility:} reduced reliance on expensive, intrusive, or failure-prone physical sensors, allowing continuous monitoring in situations where installing or maintaining direct measurement devices would be impractical.
(3) \textbf{Data completeness and enrichment:} reconstruction of variables already monitored by physical sensors to fill temporal or spatial gaps caused by intermittent or faulty measurements, and generation of derived features (such as health indices) that provide additional, physically meaningful inputs for PHM applications.

Despite their advantages, soft sensing in PHM also face several limitations. Model-driven approaches  require accurate system knowledge and reliable calibration data. 
 Developing and maintaining such models for complex assets (e.g., wind turbines, jet engines, or manufacturing equipment) is often labor-intensive and may be impractical when system behavior is nonlinear, uncertain, or poorly characterized. Furthermore, white-box models typically rely on steady-state operations and therefore struggle to adapt  to dynamic, real-world PHM scenarios, which limits their predictive accuracy during  transients or unexpected operating conditions 

 While Data-driven methods  have become popular in PHM for their ability to capture complex, nonlinear relationships from historical or operational data, their success relies on the availability of large, diverse, and high-quality datasets.  In practice, assets often undergo evolving degradation modes and operate under varying conditions, making it difficult for purely data-driven soft sensors to generalize reliably without frequent model updates.

\subsection*{Emerging and Unexplored Applications of Soft Sensing in PHM}

 Current research in PHM soft sensing is exploring several promising directions. One of the promising strategies for soft sensing in PHM is using PINNs (see Section \ref{sec:PINN})), and their graph-based counterparts PIGNNs (see Part I),  which integrate physical laws, such as conservation principles or degradation equations, directly into model architecture. This approach enables soft sensors that remain consistent with known dynamics even in data-scarce environments or under distribution shifts. Foundation models represent another promising direction. They can be prompted, through in-context learning (see Section \ref{sec:few_shot}), with structured time-series data, process descriptions, and historical logs to perform tasks like auxiliary variable selection, uncertainty-aware prediction. Multimodal (see Section 3.4) foundation models can also combine heterogeneous data sources, using language to support transparent reasoning \cite{tong2025soft}. These techniques reduce reliance on large labeled datasets by incorporating domain knowledge in natural language form. These advanced soft sensing methods have significant potential in PHM, where robustness, data efficiency, and interpretability are critical. Another promising direction is transfer learning or domain adaptation (see Section \ref{sim2real}) which allows models developed on one fleet or system to be adapted to others with minimal new data \cite{niresi2025efficientunsuperviseddomainadaptation}.

%% file: PaperII/06_Simulation_to_real_gap.tex
\newcommand{\ieno}{\textit{i}.\textit{e}.}

\begin{table}[htbp]
\centering
\belowrulesep=0pt
\aboverulesep=0pt
\renewcommand\arraystretch{2}
\begin{tabular}{@{}ccp{8cm}@{}}
\toprule
\multirow{2}{*}[-1ex]{Applications in PHM} & Type of tasks     & Fault detection; fault diagnosis; RUL prediction. \\ \cmidrule(l){2-3} 
                   & Addressed challenges & Simulation‑to‑real gap; rare failure modes\\ \midrule
\multirow{4}{*}{Requirements}    & Prior knowledge    & Physics-based models; First-principles simulations; Domain expertise\\ \cmidrule(l){2-3} 
                   & Data         & Synthetic data from simulations; limited real measurements.     \\ \cmidrule(l){2-3} 
                   & Type of bias      & Observational bias \\ \cmidrule(l){2-3} 
                   & Assumptions      & Physics-based models capture relevant degradation mechanisms; Domain gap is manageable for the chosen approach.  \\ \midrule
\multirow{2}{*}[-4ex]{Summary}    & Advantages       & Reduced dependency on labeled real-world data; Coverage of rare failure scenarios \\ \cmidrule(l){2-3} 
                   & Disadvantages     & Simulator fidelity; Complexity in implementation and validation.\\ \bottomrule
\end{tabular}
\vspace{0.5cm}
\caption{Checklist for \textit{Bridging the Simulation‑to‑Real Gap in PHM}}
\label{checklist:sim2real}
\end{table}

\subsection*{Introduction to Sim2Real domain gap}

Physics-based approaches have long been fundamental in PHM, providing interpretable, generalizable, and physically grounded models for fault detection \cite{orsagh2003prognostics,zong2016dynamic}, fault diagnosis \cite{tadina2011improved, singh2015extensive}, and prognostics~\cite{li2005gear,cubillo2016review}. By leveraging first-principles models (e.g., governing equations) and domain expertise ~\cite{daigle2011model,daigle2012model}, these methods often achieve high predictive accuracy with relatively limited data requirements. However, their effectiveness is typically confined to critical or well-understood components \cite{chao2022fusing}. In complex engineering systems, interactions among subsystems and an incomplete understanding of failure mechanisms often mean physics‑based models fail to capture real degradation dynamics \cite{ma2025bridgingrealitygapdigital}.

Despite their potential, the adoption of physics-based models remains limited in practice \cite{lei2018machinery}, due to the complexity of their development and the expertise required for accurate parameterization. Recent advances have combined physics-based models with ML to alleviate the scarcity of labeled real-world data in PHM~\cite{khan2018review}. One widely used strategy is to generate synthetic training data from physics-based simulations \cite{eklund2006using,sobie2018simulation,hagmeyer2022integration,thelen2022comprehensive}. These synthetic datasets can serve multiple purposes, such as complementing sparse real-world measurements, creating examples of rare failure modes, and providing controlled variations of operating conditions. Typically, these physics-based simulations are used to model system dynamics and degradation processes, enabling researchers to generate comprehensive datasets that would be difficult or impossible to obtain from real-world operations. 
A major challenge in leveraging synthetic data for PHM is the Simulation-to-Real gap (Sim2Real), which is the discrepancy between idealized mathematical representations (simulations) and the complexities of real-world operating conditions. Even high-fidelity simulations cannot perfectly capture real-world variability, resulting in a distribution gap between synthetic data and experimental measurements of mechanical systems \cite{gao2020fem,d2024predictive}.  Table \ref{checklist:sim2real} provides a condensed checklist for applying Sim2Real strategies in PHM, outlining the primary applications, requirements, and trade-offs.

\subsection*{Applications to PHM}

To address the Sim2Real domain gap, researchers have developed strategies that leverage both physics-based simulations and real-world data in data-driven PHM applications \cite{wang2018hybrid,nascimento2019fleet,chao2019hybrid}. These approaches aim to mitigate the inherent differences between simulated and real environments. The strategies differ in how they handle domain discrepancies and the requirements for labeled real-world data, with each approach offering distinct trade-offs between implementation complexity and data requirements.
Figure~\ref{fig:Sim2real_overview} summarizes three key strategies:

\begin{figure*}[h!]
 \centering
 \includegraphics[width=1.0\linewidth]{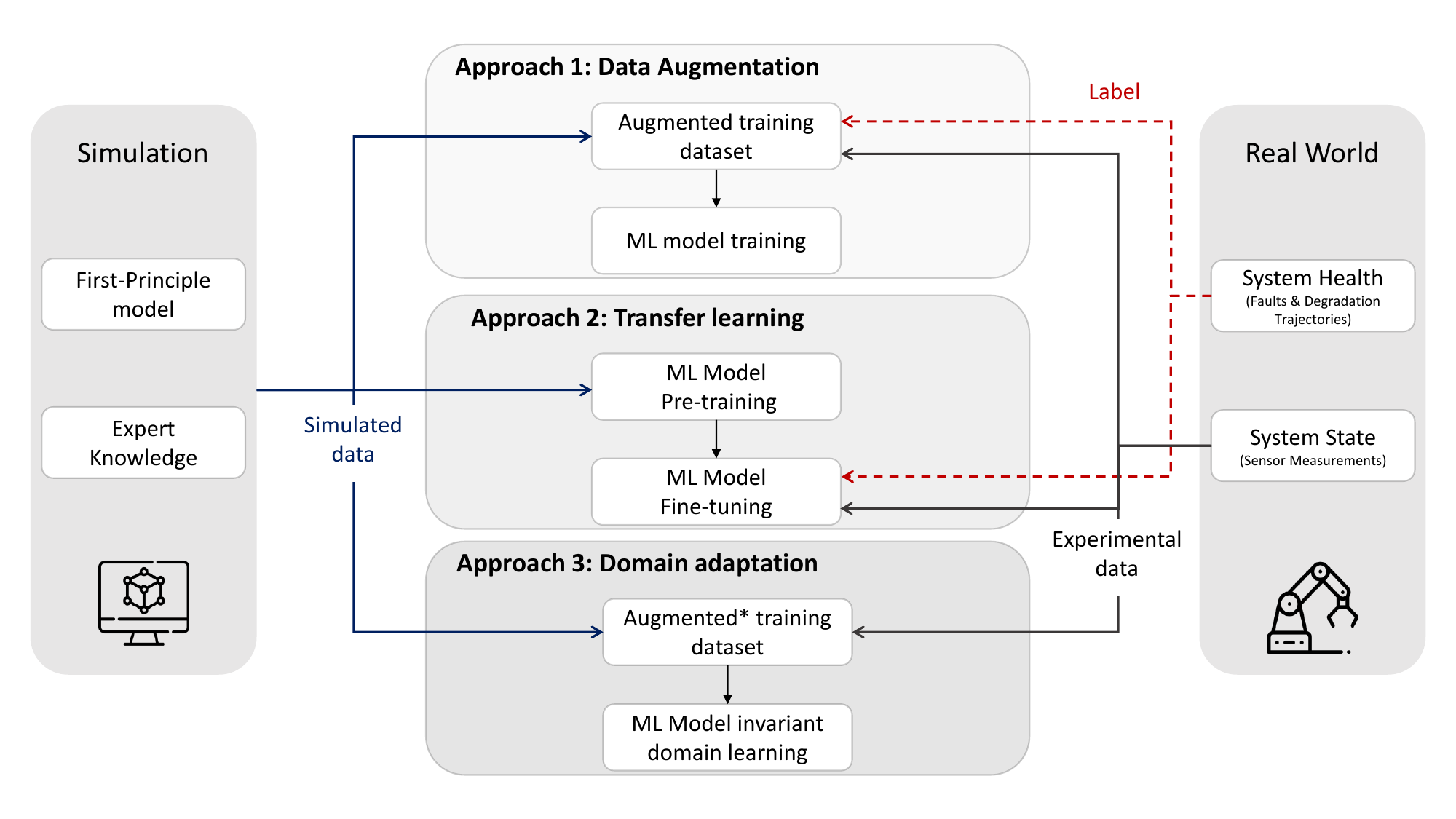}
   \vspace{-0.7cm}
  \caption{Overview of approaches for bridging the simulation-to-real gap in PHM. The figure illustrates three primary strategies: (1) Data augmentation, which enriches training data by combining simulated and experimental measurements; (2) Transfer learning, where models are pre-trained on synthetic data and fine-tuned on real-world data; and (3) Domain adaptation, which learns domain-invariant features to align synthetic and real-world distributions. Blue arrows indicate simulated data, black arrows represent experimental data, and red dashed arrows denote labels.}
  \label{fig:Sim2real_overview}
\end{figure*}

\textbf{• Approach 1: Data Augmentation} This approach enriches training datasets by combining simulated data—generated with first‑principles models across diverse degradation and operating scenarios—with experimental or in‑service measurements \cite{chao2022fusing,frank2016hybrid,huber2023physics}. By incorporating physics-informed synthetic data, this method enhances dataset diversity with additional operating conditions and degradation scenarios that might be rare or unavailable in real-world datasets. This increased diversity can lead to better generalization performance. Additionally, this approach reduces the amount of real-world data required for training. An essential condition for successful data augmentation is that the domain gap between simulation and reality should be relatively small, as this method does not explicitly address domain differences. In other words, the first-principles model must accurately capture the relevant physics of the problem.\\

\textbf{• Approach 2: Transfer Learning} This approach involves pre-training machine learning models using large-scale simulated data from first-principles simulations before fine-tuning them on a smaller set of labeled real-world measurements \cite{chakraborty2021transfer,schroder2022using,biggio2023ageing}. Similar to the data augmentation approach, the transfer learning approach improves generalization and reduces the need for large labeled datasets. 

\textbf{• Approach 3: Domain Adaptation} This method aims to align distributions between simulated and real-world data, enabling models to generalize across domains despite the absence of labeled data in the target domain \cite{liu2020domain,wang2021integrating,dong2025adapting}. Unlike transfer learning, which fine-tunes a pre-trained model using labeled data, domain adaptation learns domain-invariant features to reduce the impact of domain shifts without requiring any labels in the target domain. By improving the consistency between training and deployment domains, these methods enhance model reliability in real-world PHM scenarios.

\subsection*{Advantages and Limitations}

The integration of physics-based data into data-driven methods offers distinct advantages for PHM, such as improved generalization to unseen operating conditions while reducing the need for a large number of labeled real-world data 
\cite{sobie2018simulation,gecgel2021simulation, khan2021synthetic} -- a critical benefit given the cost, complexity, and time required to acquire run-to-failure datasets for industrial systems \cite{jan2023artificial}. 

In practice, this beneficial form of observational bias is introduced deliberately by incorporating physically-realistic synthetic data into the training process. A straightforward implementation uses high-fidelity numerical models that replicate fault or degradation behaviors, potentially covering rare or extreme operating conditions \cite{de2024remaining}. However, this can introduce a significant domain gap if the simulator does not fully mirror real-world complexities. The main limitation, however, is the scarcity of realistic degradation models; consequently, most studies generate only “healthy” synthetic data to fill under‑represented operating conditions. A second limitation is that models typically treat components in isolation, ignoring their inter‑dependencies, which further widens the gap between simulation and reality. An alternative strategy generates hybrid data by injecting simulated anomalies \cite{randall2001relationship} into healthy real-world signals, thereby preserving relevant operating conditions and reducing the domain gap between simulated and real domains.

Furthermore, generating data from high-fidelity numerical models (e.g. using the finite element method) can be computationally prohibitive. Even a single simulation of a degradation process (such as fatigue crack growth) may take several hours to run
\cite{lu2022physics,suttakul2024role}. To overcome this bottleneck, many researchers now develop computationally efficient surrogate models that approximate the behavior of the high-fidelity simulation while drastically reducing runtime \cite{torzoni2023multi}. These surrogate models serve as accurate proxies for the full-order simulations, enabling rapid analysis and prognostics without the exorbitant computation cost.

While data augmentation can significantly improve model performance, its effectiveness depends on both the fidelity of the physics-based simulations and their relevance to the specific classification or regression task. This highlights a key assumption in bridging the simulation-to-real gap: the extent to which the simulated phenomena accurately reflect  real-world degradation processes in target application. However, systematic studies assessing the maximum tolerable domain gap for each strategy remain limited.

\subsection*{Emerging Applications for Bridging the Simulation-to-Real Gap in PHM}

A promising strategy for bridging the simulation-to-real gap in PHM is the use of large-scale foundation models, neural networks pre‑trained on extensive, heterogeneous corpora that learn general representations and can be adapted to downstream tasks via fine‑tuning, few‑shot, or even zero‑shot transfer\cite{bommasani2022opportunitiesrisksfoundationmodels,zheng2024empirical}. Foundation models have shown remarkable success in different scientific domains, such as protein structure prediction \cite{abramson2024accurate}, advanced weather forecasting \cite{bodnar2024aurora,price2025probabilistic}, and their emerging applications in scientific machine learning \cite{subramanian2023towards}. The key advantages of foundation models stem from their two-phase approach. First, Pretraining, where a single large-scale neural network learns to capture relevant patterns and structural relationships from extensive, diverse datasets. Second, fine-tuning, which enables the model to leverage these learned representations to excel at new tasks and adapt to different domains with limited labeled data \cite{brown2020language}.

Recent operator‑transformer foundation models for scientific data
(e.g.\ POSEIDON \cite{herde2024poseidonefficientfoundationmodels})
show that a network can be pre‑trained on a limited set of PDEs and
later zero‑shot generalise to \emph{unseen} equations, provided the
underlying physics share structural similarities. Translating this
idea to PHM, one could pre‑train on large‑scale, multi‑condition
bearing simulations (where mechanistic models are mature) and
then adapt to new related rotary components (e.g.\ gears) by
fine‑tuning on a small amount of real data
\cite{peng2025bearllm,liu2025aerogpt}.

Beyond fine-tuning, these foundation models also offer potential for in-context learning \cite{akyurek2022learning,bai2023transformers,nejjar2024context} capabilities (see Section \ref{sec:few_shot}), where the model could analyze new degradation patterns without explicit retraining or fine-tuning them to real-world data. This would enable rapid adaptation to novel failure modes or operating conditions by simply providing relevant examples within the input context and having the model pre-trained only on synthetic data.

%% file: PaperII/07_METHOD_GENERATIVE_MODEL.tex
\begin{table}[htbp]
\centering
\belowrulesep=0pt
\aboverulesep=0pt
\renewcommand\arraystretch{2}
\begin{tabular}{@{}ccp{8cm}@{}}
\toprule
\multirow{2}{*}[-1ex]{Applications in PHM} &  Type of tasks         & Fault detection; Fault diagnosis; RUL prediction. \\ \cmidrule(l){2-3} 
                                     & Addressed challenges  & Data scarcity. \\ \midrule
\multirow{4}{*}{Requirements}        & Prior knowledge        & Governing PDEs (optional).  \\ \cmidrule(l){2-3} 
                                     &  Data                  & Enough data samples to represent the data distribution.           \\ \cmidrule(l){2-3} 
                                     & Type of bias           & Observational bias. \\ \cmidrule(l){2-3} 
                                     & Assumptions            & The underlying data distribution can be learned by the designed model.    \\ \midrule
\multirow{2}{*}[-4ex]{Summary}       & Advantages             & Generalizability and adaptability; Reduced dependency on large dataset; Rare sample generation. \\ \cmidrule(l){2-3} 
                                     & Disadvantages          &  Computational inefficiency; Synthetic-real domain gap; Bias propagation.\\ \bottomrule
\end{tabular}
\vspace{0.5cm}
\caption{Checklist of \textit{Generative Modeling}}
\label{checklist:gen}
\end{table}

\subsection*{Introduction to Generative Modeling}
Deep generative models (DGM) are neural networks designed to approximate complex, high-dimensional probability distributions using sample data~\cite{ruthotto2021introduction}. These models estimate the likelihood of observed data and generate new samples that follow the learned distribution. With their high expressiveness, DGMs effectively capture complex data patterns, making them valuable for enhancing generalizability, reducing dependence on large labeled datasets, and generating rare samples to improve model robustness. Despite challenges such as computational inefficiency, the synthetic-to-real domain gap (sec. ~\ref{sim2real}, and the risk of bias propagation, DGMs remain a powerful tool for data augmentation.

Various deep generative models have been developed to learn data distributions. Among the most well-known are generative adversarial networks (GANs)~\cite{goodfellow2014generative} achieve this by introducing two competing neural networks:  a generator, which learns to create realistic synthetic samples, and a discriminator, which distinguishes between real and generated samples. 
In contrast, Auto-encoders (AEs)~\cite{kingma2013auto} take a different approach by learning to reconstruct original inputs while compressing the data into efficient latent representations, thereby capturing the essential features of the underlying distribution. A widely used variant, variational autoencoders (VAEs), introduce a probabilistic framework for encoding and decoding, allowing the model to learn a smooth and continuous latent space that facilitates both reconstruction and generation of new samples~\cite{kingma2019introduction}. 
Flow-based generative models, such as those introduced in~\cite{rezende2015variational}, take yet another approach by explicitly modeling the data distribution through a sequence of invertible transformations.
More recently, diffusion models~\cite{sohl2015deep}, inspired by non-equilibrium thermodynamics, have gained significant attention. These models define a Markov chain that incrementally adds noise to data during training, and then learns to reverse this process, reconstructing high-fidelity samples from pure noise.

\subsection*{Physics-informed Data Generation}
Despite their effectiveness in data generation, DGMs often require large datasets and may not always align with underlying physical laws. To improve their alignment with real-world phenomena, an emerging research direction focuses on integrating physics prior knowledge into generative models by embedding  physical laws and invariance properties directly into the model architecture or training process. 

One fundamental approach, as discussed in the Sim2Real section (sec.~\ref{sim2real}), is to leverage  physics-based simulations to generate training data for generative models. This strategy helps ensure that the generated datasets retain key physical properties with controlled variations while augmenting insufficient real-world data~\cite{eklund2006using,sobie2018simulation,hagmeyer2022integration, nguyen2023physics}. 
Another strategy is to interpret or encode data in terms of physically meaningful variables, thereby improving the model’s adherence to physical principles. For instance, rather than directly predicting the wireless channel signals, Böck et al.~\cite{bock2025physics} employed a VAE to estimate signal transmission parameters, which were subsequently  used to reconstruct the channel matrix.

Beyond physics-informed data simulation and interpretation, researchers have developed methods to directly integrate physics into the training process of generative models. These approaches aim to enhance model reliability, interpretability, and generalization ability by embedding physics in different ways, ensuring consistency with the underlying laws of physics.
One prominent research direction involves incorporating physics constraints into the loss function to enforce adherence to the governing equations. This approach, pioneered by PINNs (sec.~\ref{PINNS}), uses physical laws as additional learning objectives during training. This framework has been extended to DGMs, for example Zhou et al.~\cite{zhou2022physics} proposed a physics-informed GAN for system reliability assessment, which fuses limited measurement data with underlying mathematical models that  aggregates the system state probability in evolution of time. Their method demonstrated  effectiveness on a simulated dual-processor computing system, highlighting its potential in scenarios  where real-world failure data is scarce.
In addition to physics-informed loss functions, researchers have explored integrating physical principles directly into the architectural design of generative models, such as graph neural networks (GNNs) for physical reasoning~\cite{battaglia2016interaction}, Group equivariant Convolutional
Neural Networks (G-CNNs) to model physical symmetry~\cite{cohen_group_2016} and Hamiltonian Neural Networks (HNNs)~\cite{greydanus2019hamiltonian}. 
With the advancement of diffusion models, Bastek et al.~\cite{bastek2024physics} introduced Physics-Informed Diffusion Models (PIDMs), which integrate physical constraints into the training of denoising diffusion models, enabling the generation of samples that comply with governing PDE equations. Along similar lines, Qiu et al.~\cite{qiu2024pi} applied PIDMs for fuild dynamics, proposing Pi-fusion -- a method that integrates a physics-informed guidance sampling strategy into the diffusion process to capture quasi-periodical patterns in fluid motion, thereby demonstrating the potential of PIDMs for complex physical systems. Furthermore, Rissanen et al.~\cite{rissanen2022generative} presented a generative modeling framework inspired by the physics of heat dissipation, leveraging the inverse process of heat diffusion to generate data that closely matches  the true distribution.

\subsection*{Generative modeling for PHM}

\begin{figure}
    \centering
    \includegraphics[width=1.0\linewidth]{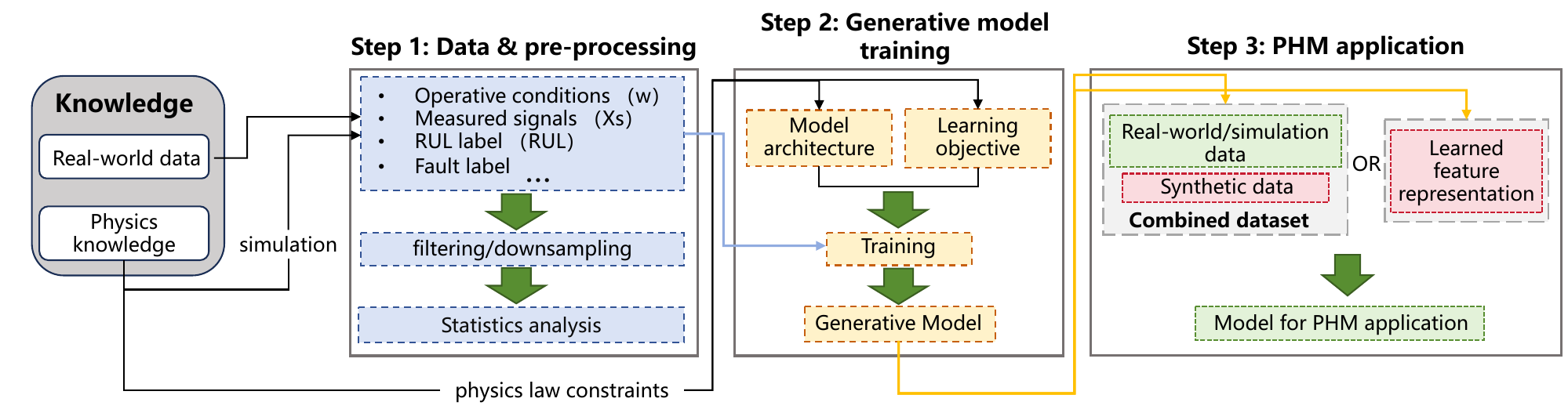}
    \caption{General framework of generative modeling for PHM applications. The process consists of three main stages: (1) Data collection and pre-processing: collect real-world data, optionally augmented with physics-based simulated data, followed by data pre-processing; (2) Generative model training: train generative models to learn the underlying data distribution of input data, incorporating physical priors where applicable; and (3) PHM application: use the trained model to generate synthetic data, extract informative features to support downstream tasks such as fault diagnosis, anomaly detection, and remaining useful life prediction.}
    \label{fig:generative}
\end{figure}

A major challenge in PHM applications is the scarcity of labeled training data in real-world industrial settings, as discussed in the previous subsection on simulation-to-real (sec.~\ref{sim2real}) and in the section of scaling beyond single systems (sec.~\ref{scaling}). 
In response to this challenge, generative modeling has become an increasingly popular strategy for  learning data distributions and generating realistic synthetic data in a flexible and efficient manner, as illustrated in the general framework shown in Figure \ref{fig:generative}.

In scenarios characterized by data scarcity or imbalance, generative models are widely used to augment datasets and compensate for missing or underrepresented distributions. For instance, VAEs can learn the underlying data distribution and generate additional samples for minority  classes, resulting in more diverse and balanced datasets. This approach has proven effective for fault diagnosis of bearing rolling ~\cite{liu2023interpretable, zhao2022normalized, tian2024novel} and multi-phase flow process~\cite{tian2024novel} under imbalanced conditions. It has also been applied for RUL prediction of turbofan aircraft engines~\cite{costa2025few}. Beyond simply enriching data distribution, generative models are increasingly applied  to synthesize rare faulty data, which is typically difficult  to obtain  in industrial systems. Hong et al.~\cite{hong20231d}, for instance,  combined VAEs with GANs to augment noise and vibration data, enhancing  the representation of rare fault conditions and improving the accuracy of bearing fault diagnosis.  Furthermore, generative models have been applied to address scenarios involving unseen fault conditions by simulating novel fault types, thereby improving the generalization capability of diagnostic models. For instance, Rombach et al.\cite{rombach2023controlled} utilized GANs to generate synthetic data representing previously unseen fault types under specific operating conditions, thereby enabling partial open-set domain adaptation. Sun et al.~\cite{sun2025unseen} deployed diffusion model to generate visual unseen anomalies for the unsupervised anomaly detection task. These  approaches effectively addresses the challenge of limited real-world fault data by enriching the dataset with diverse fault scenarios.

Beyond data augmentation, generative models are increasingly applied to learn the underlying data distributions during the generative training, enabling the extraction of compact and informative representations. For instance, Ren et al.~\cite{ren2022vibration} introduced a multivariate invertible deep probabilistic framework to uncover latent patterns in gearbox fault diagnosis. By distilling  essential low-dimensional features from high-dimensional sensor data, their method achieved more robust and accurate fault classification. In addition to feature learning, DGMs are leveraged to model the normal behavior of complex systems. This approach is particularly valuable for fault detection in scenarios where faulty data is scarce or unavailable. By learning the distribution of normal system behavior, these models can identify deviations as potential faults. For instance, VAE-based generative models have been used for time series reconstruction in wind turbine anomaly detection, successfully identifying deviations from expected signal patterns ~\cite{luo2023multi}. Hu et al.~\cite{hu2025classifier} proposed a diffusion-based weakly-supervised approach for deriving health indicators of rotating machines, by training a diffusion model to generate normal samples. Similarly, Zhu et al.\cite{zhu2019novel} combined the temporal modeling capabilities of LSTM networks with the generative power of GANs, enabling the capture of complex temporal dependencies and dynamic behaviors. The fused model design significantly enhances fault detection by combining the generator's reconstruction error (the residual) and the discriminator's classification loss.

Physic-informed data generation has been increasingly applied in PHM to achieve more interpretable and reliable model performance by integrating known physical principles as constraints within generative models. For example, Gao et al.\cite{gao2020fem} introduced a GAN-based approach to generate synthetic training data that complements physics-based simulation outputs. By increasing data diversity, their method improves the generalization ability of models for bearing fault detection. Expanding on this idea, Nguyen et al.\cite{nguyen2023physics} proposed a fuzzy GAN that embeds physical degradation models for RUL prediction to produce degradation trajectories that accurately 
 mimic real-world system  behavior, resulting in more robust RUL prediction for bearing systems and C-MAPSS dataset. To further integrate physical principles into the generation process, Fan et al.\cite{fang2021fault} first performed feature extraction to capture relevant physical characteristics, which were then used as reference vectors in a GAN-based model. This approach facilitated the generation of synthetic samples tailored to infrared images, showcasing how physics-informed feature extraction can improve diagnostic accuracy. 
Another widely adopted strategy in PHM  involves incorporating physics-based constraints directly into the loss functions of learning models. 

Xu et al.\cite{xu2022physics}, for instance, developed a physics-informed dynamic depth autoencoder that integrates battery state and capacity equations as penalty terms in its loss function. This design improves both the accuracy and interpretability of the lithium battery secondary variables, capacity, and state-of-health prediction results. Similarly, Xiong et al.\cite{xiong2023controlled} introduced the Controlled Physics-Informed GAN (CPI-GAN) for RUL prediction of turbofan engines under conditions of limited time-to-failure data. CPI-GAN  generates synthetic trajectories of operating conditions that integrate degradation by incorporating five basic physics-based constraints as controllable terms within the loss function to enforce the generator. These constraints enforce consistency between the generator’s outputs and the underlying physical laws, thereby enhancing both the performance and reliability of the RUL predictions.
In terms of architecture design, PG-GAN~\cite{gao2023physics} allows
incorporating physical information within the architecture and training process of GAN. This enhanced GAN-based architecture uses a VAE to structure its generator's input, an ESO to provide its discriminator with real physical data, and a physical loss function to improve the generator's training. Sanchez et al.~\cite{sanchez2025addressing} augment sensor data using Physically Informed Echo State Networks
(ESNs). Their approach integrates
domain-specific physical knowledge into the learning process to generate surrogate time-amplitude
signals from the Power Spectral Density of the data and is tested on the real-world jet fan dataset.  
A summary of the application contexts for generative modeling is provided in Table \ref{checklist:gen}.

\subsection*{Advantages and disadvantages of generative modeling in PHM}

Generative models are playing an increasingly important role in PHM applications by addressing critical data limitations inherent to the field, including the scarcity of faulty samples, incomplete coverage of possible fault types, and the limited availability of complete time-to-failure trajectories under diverse operating conditions.
A key advantage of these models is their ability to generate synthetic data to overcome these limitations. By sampling from low-density regions of the learned data distribution, they are able to generate more data for rare or underrepresented failure scenarios. This not only improves the representativeness of the dataset but also helps to balance heavily skewed datasets.
Specifically, these models can generate new examples of rare failures to address the scarcity of faulty samples, and they can extrapolate from the learned distribution to generate unseen fault types, tackling the problem of incomplete fault coverage. For prognostics, these models can synthesize entire time-to-failure trajectories, providing the complete degradation data needed to train robust models for RUL prediction.
Beyond data augmentation, generative models excel at learning robust representations of normal system behavior.  Through training on data generation tasks, they learn to capture the intrinsic patterns of healthy operation across all conditions. This directly addresses the challenge of highly varying operating conditions, as the models can learn to distinguish between changes due to system operation and those indicating actual degradation. This capability greatly improves the generalization ability of the fault detection system, making it highly effective in detecting anomalies or deviations that may indicate potential faults.

However, applying generative models in PHM applications also comes with challenges. These models often require high computational resources and large amounts of high-quality training data, which may not always be available in real-world scenarios. Additionally, their complexity makes them difficult to interpret, which can be a drawback in safety-critical applications. Without proper regularization and sufficient data, these models also risk overfitting to the observational biases within the limited dataset, leading to biased learned distributions that fail to generalize to new conditions. In addition, the synthetic samples produced by these models may not always align with real-world physical laws, leading to potential inaccuracies. Despite these limitations, generative models remain an active area of research for improving predictive maintenance and fault detection.

\subsection*{Unexplored or Emerging Applications}

While generative models have demonstrated significant potential in PHM, many applications remain unexplored or are only beginning to emerge. One promising direction is the integration of physics-informed constraints into generative models. By embedding prior physical knowledge,  these models can generate data that adheres to  the fundamental  principles governing  industrial systems, thereby enhancing  interpretability, trustworthiness and reliability in PHM applications. Another emerging area of development is the data-efficient generative modeling, which aims to improve the  effectiveness of these models  in real-world PHM applications in scenarios characterized by limited or imbalanced data and constrained computational resources.

%% file: PaperII/12_METHOD_DATA_FUSION.tex

\begin{table}[ht!]
\centering
\belowrulesep=0pt
\aboverulesep=0pt
\renewcommand\arraystretch{2}
\begin{tabular}{@{}ccp{8cm}@{}}
\toprule
\multirow{2}{*}[-1ex]{Applications in PHM} &  Type of tasks         & RUL prediction; Fault detection; Fault diagnosis.  \\ \cmidrule(l){2-3} 
                                     & Addressed challenges  &  Uni-modal bias; Incomplete health representation; Degradation modeling. \\ \midrule
\multirow{4}{*}{Requirements}        & Prior knowledge        &  Sensor models; Modality-specific representation. \\ \cmidrule(l){2-3} 
                                     & Data                  &  Heterogeneous or homogeneous data. \\ \cmidrule(l){2-3} 
                                     & Type of bias           & Observational bias. \\ \cmidrule(l){2-3} 
                                     & Assumptions            & Different modalities are interconnected with shared complementary
information.   \\ \midrule
\multirow{2}{*}[-4ex]{Summary}       & Advantages             & Cross-modal data compensate for individual weaknesses and improve overall system accuracy and reliability.  \\ \cmidrule(l){2-3} 
                                     & Disadvantages          & Inherent differences in the statistical properties of each modality and the complex nonlinear relationships among their raw representations. \\ \bottomrule
\end{tabular}
\vspace{0.5cm}
\caption{Checklist of \textit{Data Fusion}}
\label{checklist:fusion}
\end{table}

\subsection*{Introduction to Data Fusion}
Data fusion refers to the integration of information from diverse sources and modalities into a unified representation~\cite{khaleghi2013multisensor, MENG2020115}. 
By leveraging cross-domain data, it can capture different partial information about the same system condition and compensate for the limitations of individual sources, thereby improving the overall accuracy, robustness, and reliability of downstream algorithms~\cite{jardine2006review, Jose2023}.
The ability to integrate complementary information, reduce uncertainty, and improve decision-making has driven the widespread adoption of data fusion techniques across various domains.  
Notable applications include autonomous vehicles~\cite{yeong2021sensor,SuperFusion}, computer vision~\cite{chen2023futr3d,dong2023jras}, and wearable health monitoring systems~\cite{king2017application}. 
In PHM, the increasing complexity of industrial systems makes it challenging for uni-modal data to accurately describe overall degradation processes. 
As a result, 
data fusion techniques enable the integration of additional heterogeneous data streams (e.g.,  temperature, pressure, vibration) to support fault detection, diagnostics and prognostics in systems like aircraft engines~\cite{6678166, sarkar2013multi},
gearboxes~\cite{xia2023novel}, and electric vehicles~\cite{el2024toward}.

From a PIML perspective, 
multimodal data fusion inherently introduces observational bias, not merely as noise or artifacts, but as a set of constraints imposed by the selection, representation, and sampling of input data~\cite{liang2024foundations}.
In practice, the choice of sensors, included modalities, spatial configuration, and underlying measurement principles collectively determine which aspects of the physical system are emphasized or omitted in the fused representation~\cite{grade}.
Rather than being a direct reflection of the physical system, the integrated data is shaped by the sensitivities and limitations of each modality, creating a composite representation defined by both the complementarity and incompleteness of the data sources.
In PIML, these observational biases influence 
model learning by shaping which physical laws or latent variables can be inferred from the available observations~\cite{Jose2023, SUN2024102394}. While such biases can ensure that models remain grounded in sensor-accessible, physically relevant information, they also risk obscuring or distorting less observable phenomena. Explicitly acknowledging and managing this bias is therefore essential for PIML as it enables models to leverage the strengths of data fusion for robust and interpretable inference, while mitigating the risk of spurious correlations or information gaps that may compromise physical reasoning and prediction.

Data fusion methodologies are commonly categorized by the characteristics of data sources involved: heterogeneous and homogeneous fusion. 
Heterogeneous data fusion combines information from diverse modalities or sensor types. In PHM, commonly used data include vibration, acoustic, temperature, pressure, speed, electric current, and voltage signals, with visual and text data being increasingly used in recent years~\cite{Jose2023, SUN2024102394}.
However, significant challenges remain due to inherent variations in data quality, structure, and representation formats~\cite{liang2024foundations}.
In contrast, homogeneous data fusion integrates data from multiple sources of the same modality.
For instance, multi-camera systems positioned at \textit{different locations} capture complementary visual perspectives of the same scene, supporting applications such as building inspection~\cite{WANG2023104810, xu_exploiting_2025} and rail defect detection~\cite{hu20243d, bhavanasi2025enhanced}.

\begin{figure}
    \centering
    \includegraphics[width=\linewidth]{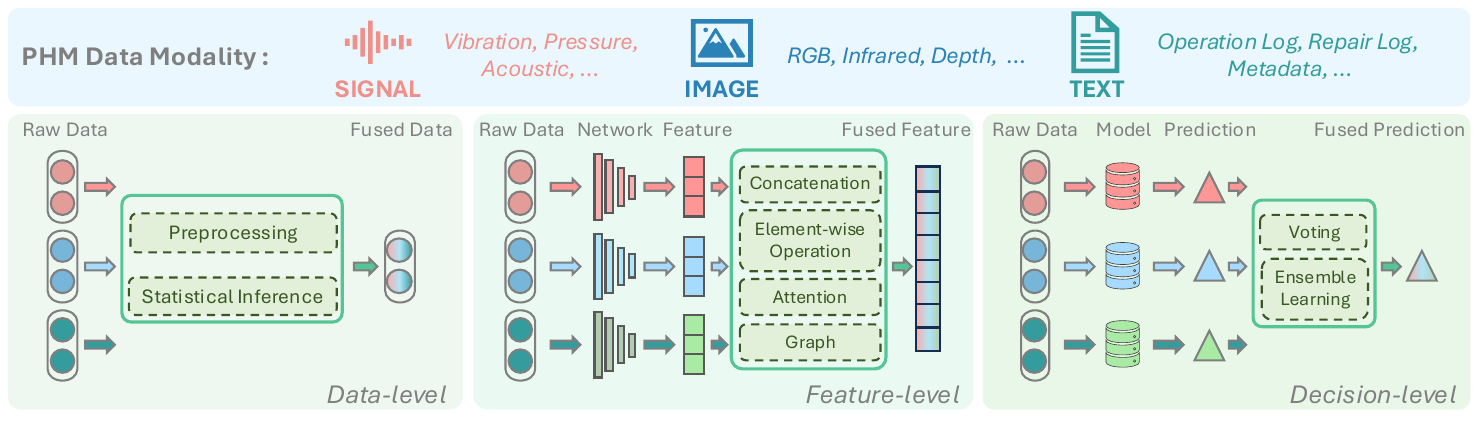}
    \caption{Overview of data fusion strategies in PHM applications. This includes three primary fusion stages: (1) \textbf{Data-level fusion} combines raw sensor data through preprocessing or statistical inference to generate integrated data representations; (2) \textbf{Feature-level fusion} extracts modality-specific features and integrates them using advanced techniques; and (3) \textbf{Decision-level fusion} merges predictions from individual models trained on different modalities using strategies like voting or ensemble learning.}
    \label{fig:datafusion}
\end{figure}

Recent advancements in multimodal machine learning have broadened the scope of data fusion in two key directions: (1) the development of novel neural representations by integrating common modalities, such as combining infrared images with vibration signals for gearbox diagnosis~\cite{10874640}; and (2) the incorporation of previously underutilized modalities such as natural language and semantic metadata, for example, integrating technical documents and maintenance logs into diagnostic models for hydrogenerators~\cite{yin2024survey, JOSE2024124603}. 
To systematically categorize recent data fusion methodologies, we adopt a common classification framework based on three fusion stages~\cite{luo2002multisensor, MENG2020115}:
\textbf{data-level fusion}, \textbf{feature-level fusion}, and \textbf{decision-level fusion}, as illustrated in Figure~\ref{fig:datafusion}. Each level presents distinct trade-offs in terms of complexity, interpretability, and performance.

\textbf{Data-level fusion}
focuses on the direct combination of raw data streams to estimate system states or conditions~\cite{castanedo, MENG2020115, 10.1145/3649447}. 
Classic approaches are typically classified into two complementary groups: preprocessing techniques and statistical inference models.
Preprocessing techniques, such as image processing and signal filtering~\cite{LI2017100, pajares2004wavelet, rubinstein2010dictionaries}, operate directly on raw data to achieve spatial, temporal, or statistical alignment of multimodal inputs. 
These methods preserve the \textit{original data format} while enhancing accuracy and minimizing noise, thereby ensuring that downstream models operate on well-calibrated, high-quality input data. 
In contrast, statistical inference methods, including the Kalman Filter, Extended Kalman Filter, Unscented Kalman Filter, and Particle Filter, recursively integrate raw sensor observations to estimate the state of dynamic systems (\textit{prediction} stage) while compensating for measurement and process noise (\textit{correction} stage)~\cite{urrea2021kalman}. Further details on the use of Kalman filtering for state transitions and sequential modeling are provided in the State-Space Models section of Part I.

\textbf{Feature-level fusion}
extracts high-dimensional representations from preprocessed data and performs cross-modal integration to construct more discriminative feature representations~\cite{10770236}. 
In the context of feature-level fusion, traditional feature extraction techniques used in ML-based PHM applications, such as Fourier and wavelet transforms~\cite{pajares2004wavelet}, are typically designed for a single modality. As a result, they are generally inadequate for fusing multimodal inputs that differ in sampling rates, physical units, or data formats (e.g., time-series signals, images, or tabular measurements)~\cite{8269806,AYMAN2025116589}.
In contrast, recent approaches leverage deep learning architectures that automatically \textit{learn} relevant features from data. These architectures are typically applied independently for each modality to effectively capture modality-specific information~\cite{10.1145/3649447, 10.1162/neco_a_01273}.
To fuse these modality-specific representations, various fusion strategies have been proposed. Basic methods, like concatenation or element-wise operations on learned features, are widely used~\cite{yin2025remaining, 9259076} but often fail to capture inter-modality correlations, potentially retaining redundant information.
Recent studies incorporate techniques such as Canonical Correlation Analysis or attention mechanisms to emphasize complementary information while suppressing irrelevant or overlapping features  across modalities~\cite{10049211, 10.1145/3649447, 10770236}.
Moreover, graph-based models, including Graph Neural Networks, Graph Convolutional Networks, and Graph Attention Networks, have emerged, enabling the modeling of complex interdependencies and contextual relationships within multimodal data~\cite{10.1145/3649447, theiler2025heterogeneousgraphneuralnetworks}.

\textbf{Decision-level fusion} combines predictions from multiple independent single-modality models to reach a final decision~\cite{10770236, PU2023107095}.
Each modality-specific model generates predictions based on its respective measurements and algorithms, thereby offering a high degree of flexibility in accommodating heterogeneous data characteristics and maintaining robustness even in the presence of sensor noise or failure~\cite{KIBRETE2024114658}.
These methods are typically categorized into statistical approaches and learning-based techniques. Statistical methods use probabilistic or rule-based frameworks, such as weighted voting~\cite{LI2020106752}, Dempster–Shafer theory~\cite{JI2020108129}, and Bayesian decision theory~\cite{XU2012816}, to reconcile uncertainty and inconsistent decisions from different modality-specific models.
In contrast, learning-based methods often adopt ensemble learning strategies, where independent sub-models are trained on modality-specific data to estimate predictive confidence or contribution weights. The final decision is then obtained by aggregating these outputs, typically through confidence-weighted averaging or voting, to improve overall reliability and accuracy~\cite{ZHONG201899, WANG2018112}.

\subsection*{Application of Data Fusion in PHM}
Benefiting from advances in machine learning and the increasing availability of multimodal data, data fusion has become a critical enabler in PHM, driving advancements in fault detection, diagnosis and prognostic through the integration of multi-source and multi-modal information~\cite{AYMAN2025116589, Jose2023}. 
A concise overview of representative application contexts, addressed challenges, and typical requirements for data fusion in PHM is provided in Table~\ref{checklist:fusion}.
At the data level, \textbf{preprocessing techniques} serve as a preliminary step to ensure high-quality, well-calibrated input data for subsequent feature- or decision-level fusion.
For example, frequency-domain transformations and cosine similarity are commonly applied to align and merge homogeneous vibration signals collected from multiple sensor positions. The fused signals are then fed into feature-level architectures for fault detection, either with CNNs combined with acoustic modalities in compressor blades~\cite{9771315}, or independently into neural networks for rotating machinery~\cite{luwei2018integrated}.
Beyond fault detection, a common data-level application in PHM is the construction of \textbf{health indices (HIs)}.
HIs are typically generated by fusing multiple sensor measurements to support degradation modeling and prognostics.
Various approaches have been proposed for HI construction, ranging from quadratic programming-based fusion~\cite{6496166, 6902828}, genetic optimization~\cite{8664723}, kernel-based fusion~\cite{7974789}, and more recently, deep neural networks that predict HIs by capturing nonlinear relationships among multiple sensor signals~\cite{10043638}.

In contrast to preprocessing techniques, \textbf{statistical data-level fusion} approaches, particularly the Kalman Filter and its variants, have been widely adopted in PHM.
By leveraging probabilistic state estimation and recursive updates,
KF-based methods combine established system models with measurements from multiple modalities to enable prognostics~\cite{7358146, LI2023109269}. However, traditional KF applications are often limited by their reliance on linear system assumptions and the requirement for precise a priori models, such as system dynamics and noise characteristics.
Recent advancements have focused on improving the adaptability and robustness of KF-based methods through integration with data-driven techniques for RUL prediction~\cite{9477442, 9758819}. While the KF assumes linear system dynamics and Gaussian noise~\cite{kalman1960new}, the Particle Filter (PF) employs sequential Monte Carlo sampling to handle highly nonlinear and non-Gaussian systems~\cite{gordon1993novel}.
For example, PF has been combined with SVM trained on fused HIs, where entropy-based fusion was employed to fuse the prognostic outputs from  PF over multiple time steps~\cite{9477442}. Similarly, Lee et al.~\cite{9758819} utilized an EKF to fuse vibration and pressure signals, which were subsequently used to train a BiLSTM network for RUL prediction.
These integrations leverage interpretability of model-based filtering and the adaptability of data-driven learning to capture nonlinear degradation patterns and complex system dynamics under time-varying conditions~\cite{Feng31122023}.

At the \textbf{feature-level}, most PHM applications have focused on \textbf{signal-based} modalities, reflecting the prevalence of vibration, acoustic, and other time-series measurements in condition monitoring. 
In this setting, features are first extracted from each modality, either through handcrafted signal processing or through learnable transformations, before integration. Common architectures include Fully Connected Neural Networks for learning nonlinear feature representations~\cite{1302346, tian2012artificial}, CNNs for capturing spatial patterns~\cite{sateesh2016deep, 9404310}, and RNNs, especially LSTM variants, for modeling temporal dependencies~\cite{7998311, 9089252}. 
Recent advances have extended feature-level fusion to increasingly heterogeneous modalities, broadening the scope of PHM applications through the integration of \textbf{image and text} data.
For example, in the visual domain, Cordoni et al.~\cite{CORDONI2022104729} 
concatenated pretrained VGG features extracted from thermal images with structured data (e.g., internal temperature, power consumption) through an autoencoder, forming a unified feature representation for refrigerator condition monitoring.
To address feature distribution misalignment in cross-modal fusion, Zhang et al.~\cite{ZHANG2024108236} combined separately extracted features from thermal images and vibration signals through fully connected layers, jointly optimizing classification, maximum mean discrepancy, and cross-modal consistency losses for gearbox diagnostics.
Similarly, the DFINet architecture incorporates a feature interaction module to enable cross-domain feature learning and a global fusion module that adaptively integrates shallow and deep representations from vibration, infrared, and spectral data for rotating machinery diagnosis~\cite{MIAO2024109795}.
For electrical power equipment inspection, capsule networks have been employed to fuse RGB, infrared, and ultraviolet imagery, leveraging vector-based features to preserve spatial relationships between low- and high-level features and jointly capture leakage, temperature, and other physical information~\cite{li2019image}.

\textbf{Text-based feature fusion} has gained increasing attention by integrating features from unstructured  textual sources, such as maintenance logs, condition reports, and technical documents, with those from other sensing modalities, including time-series data and images.
For example, Wang et al.~\cite{wang2025diagllm} formulated fault diagnosis as a \textit{visual question answering} task, employing large language models to fuse textual expert knowledge with envelope spectrum images constructed from vibration signals through the  Hilbert transform and FFT. 
In this framework, visual features from envelope spectrum images are projected into the LLM embedding space and concatenated with text embeddings, enabling multimodal reasoning that improves bearing fault diagnosis accuracy, particularly in limited data scenarios~\cite{wang2025diagllm}.
Beyond this, domain-adapted LLMs have been coupled with condition monitoring data to extract contextual insights from maintenance logs in hydropower generators~\cite{JOSE2024124603}. In this approach, the models are fine-tuned with domain-specific knowledge and fused narrative fault descriptions with sensor-derived operational parameters to enhance situational awareness. Similarly, Mandelli et al.~\cite{mandelli2022model} integrated natural language processing with model-based system engineering to fuse unstructured nuclear plant reports with structured system models, thereby linking causal and health-related information directly to component functions and dependencies.

GNNs have also emerged as a powerful framework for \textbf{feature-level fusion} in PHM.
A representative class of approaches is spatial-temporal fusion architectures, which integrate graph-based physical topology (e.g., sensor/module relationships) with temporal sensor data. In such frameworks, GNNs model the spatial structure, while LSTM- or attention-based encoders capture temporal dependencies. 
This has been demonstrated, for instance, in turbofan engine RUL prediction using hierarchical sensor–module graphs with adaptive cross-graph fusion~\cite{HGNN-ACGF}, as well as in hierarchical attention graph convolution with self-attention pooling for selective sensor integration~\cite{HAGCN}.
Building on this,  heterogeneous graph fusion methods explicitly represent complex system interactions by encoding diverse node and edge attributes.
Examples include multiplex aggregation across spatial and temporal meta-paths to fuse heterogeneous bearing vibration patterns~\cite{MAHGNN}, as well as heterogeneous graph attention mechanisms that integrate both intra-domain and inter-domain dependencies (e.g., hydraulic–electrical subsystem couplings) with distinct temporal dynamics in hydropower plants~\cite{Theiler_2024, theiler2025heterogeneousgraphneuralnetworks}. 
In addition, hierarchical feature fusion techniques capture multi-scale sensor correlations through multi-level attention mechanisms, such as node-, subgraph-, and cyclical-level modeling in aircraft engines~\cite{MAGF-SPA}, or adaptive weighting strategies applied to micrograph sequences~\cite{LOGO}.
Beyond this, several methods augment GNN-based fusion with \textbf{physics-informed or statistical regularization}, such as embedding autoregressive moving-average regression within GCNs or incorporating physics-based loss terms to prioritize timely fault prediction~\cite{GCN-ARMA}. Other approaches compute performance deterioration indices by fusing residuals between predicted and measured heterogeneous signals in hydropower turbines~\cite{FTU-HBM}. These enhancements improve both the robustness and interpretability of PHM models. A more detailed introduction to GNN architectures and related PHM applications is provided in the Graph Neural Networks section of Part I.

\subsection*{Advantages and Limitations of Data Fusion Approaches}
Data fusion techniques significantly enhance PHM systems by effectively integrating  multi-source and multimodal information, thereby improving the accuracy, robustness, and adaptability of detection, diagnostic, and prognostic models. By combining diverse data sources, fusion methods provide a more comprehensive and complementary representation of system health and degradation process, capturing failure patterns that may be overlooked by uni-modal approaches. 
From a Physics-Informed Machine Learning perspective, incorporating complementary information from different modalities reduces observational bias and aligns learned representations more closely with underlying physical processes. This enriched, physically consistent view leads to more meaningful insights and improves the generalization of health monitoring models across varying operational scenarios.

A key advantage of data fusion lies in increased robustness and fault tolerance, achieved through redundancy and complementary information. 
This mitigates the impact of noisy, incomplete, or missing data, which ensures greater reliability and continuity under potentially adverse conditions.
Moreover, data fusion enables the modeling of complex spatial, temporal, and structural dependencies often overlooked by single-source methods  through advanced techniques like Transformers and GNNs, which are capable of capturing interactions across modalities and components.
An additional benefit is improved scalability and adaptability through hierarchical and hybrid fusion architectures, which support multi-level integration and dynamic reconfiguration in response to changing system states.
Finally, by incorporating heterogeneous modalities such as text and image data, data fusion broadens the informational scope of PHM systems. This allows for context-aware diagnostics and more interpretable predictions by combining quantitative sensor measurements with unstructured expert knowledge or visual evidence.

Despite its advantages, several challenges  still limit the practical deployment and effectiveness of data fusion in PHM systems.
From a data perspective, heterogeneity remains a significant obstacle.
Sensor data from different modalities often vary in sampling rates, formats, temporal alignment, and noise characteristics. Integrating such heterogeneous streams requires extensive preprocessing and complex architectures for effective fusion.  
From a methodological perspective, a critical limitation lies in the limited interpretability of many fusion models, particularly those based on deep learning. These models often operate as black boxes, making it difficult to  validate system behavior or capture cross-modal correlations effectively. When interactions between modalities are ignored or insufficiently modeled, the resulting fused representation may not  leverage the richness of multi-source data.

\subsection*{Unexplored and Emerging Applications in PHM}
Modern data fusion in PHM is no longer limited to to traditional physical sensors such as accelerometers, strain gauges, or temperature probes. Increasingly, it incorporates \textbf{non-sensor data sources}, including maintenance records, inspection reports, fault annotations, and other semantic or contextual metadata, that provide complementary operational and historical insights.

Although the application of LLMs in PHM is still at an early-stage, aforementioned studies demonstrate their potential for \textbf{multimodal fusion}, where unstructured text is combined with structured numerical or visual sensor data. In these settings, the \textbf{text modality} typically encodes expert knowledge, historical fault cases, or procedural descriptions, whereas the \textbf{sensor modality} may consist of time-series measurements or derived representations such as envelope spectrum images. By mapping these modalities into a shared embedding space, LLMs can perform context-aware reasoning that links descriptive knowledge to observed machine states.

Further research aims to integrate 3D reconstruction (e.g., photogrammetry or neural radiance fields) and advanced computer vision (e.g., infrared anomaly detection) to construct digital twins or inspect visual degradation with high precision.
These hybrid approaches bridge the gap between physical sensor data and contextual, human-generated knowledge, advancing PHM through enhanced interpretability and hybrid physical-contextual insights.
Importantly, these advancements also highlight the need for models that can generalize across diverse operating conditions and data domains, a challenge that is increasingly central to the next stage of PHM research.

%% file: PaperII/08_METHOD_REINFORCEMENT_LEARNING.tex
\begin{table}[ht!]
\centering
\belowrulesep=0pt
\aboverulesep=0pt
\renewcommand\arraystretch{2}
\begin{tabular}{@{}ccp{8cm}@{}}
\toprule
\multirow{2}{*}[-1ex]{Applications in PHM} &  Type of tasks         & Prognostics; Fault detection; Prescriptive Operations \\ \cmidrule(l){2-3} 
                                     & Addressed challenges  & Complex decision making tasks with long-term objectives and scarce supervision\\ \midrule
\multirow{4}{*}{Requirements}        & Prior knowledge        & Environment interaction (potentially offline/simulated).  \\ \cmidrule(l){2-3} 
                                     &  Data                  & Depending on the task, many interactions with the environment are needed to obtain a robust internal representation of the task and learn a policy that accumulates satisfactory reward levels          \\ \cmidrule(l){2-3} 
                                     & Type of bias           & Learning, Inductive and Observational bias. \\ \cmidrule(l){2-3} 
                                     & Assumptions            & The environment's dynamics can be learned by the agent through observations and interactions; Sparsity of the reward signal can be mitigated  \\ \midrule
\multirow{2}{*}[-4ex]{Summary}       & Advantages             & Multi-objective optimization; 
Sequential decision-making; Integration with simulation environments and digital twins for safe policy development \\ \cmidrule(l){2-3} 
                                     & Disadvantages          & Data inefficiency; Hard to train; Hard to define a good reward signal; Scalability \\ \bottomrule
\end{tabular}
\vspace{0.5cm}
\caption{Checklist of \textit{Reinforcement Learning}}
\label{checklist:RL}
\end{table}

\subsection*{Introduction to Reinforcement Learning}
Reinforcement Learning (RL) is a branch of machine learning  in which an agent learns to make decisions by interacting with its environment \cite{sutton1998reinforcement}. The agent takes actions and receives feedback in the form of rewards or penalties, using this information to improve its strategy over time. The goal is to maximize cumulative reward by discovering optimal policies through trial and error.  Unlike supervised learning, which depends  on labeled examples, RL agents learn autonomously  from feedback signals received as a result of their actions. 
This approach is especially well-suited  for sequential decision-making problems where the optimal strategy cannot be predefined \cite{mnih2015human}. Recent advances in deep reinforcement learning have demonstrated impressive capabilities across diverse domains \cite{tang2025deep}, including  discovering a novel matrix multiplication method \cite{fawzi2022discovering}, mastering complex strategy games \cite{alphazero,tian2023multi},  controlling robotic systems \cite{han2024lifelike}, and advancing  scientific discovery \cite{esteso2023reinforcement,jansen2024discoveryworld}.  In the context of PHM, this ability to learn adaptive policies for complex, sequential decisions is particularly critical. While traditional PHM techniques focus on passive prediction, effective asset management requires translating these predictions into active decisions that directly impact the physical system. RL \textbf{closes this loop} by enabling agents to interact directly with their environment, take actions, and continuously refine maintenance and operational strategies in response to real-world feedback \cite{lee2023deep,zhang2025condition}.

\subsection*{Applications in PHM}

PHM is evolving from a focus on predicting system conditions and remaining useful life to a more proactive paradigm: prescriptive operation and maintenance \cite{lee2024stochastic,liu2024maintenance}. This emerging direction not only anticipates failures but also seeks to optimize system operations in real-time data, taking actions that extend asset life, reduce costs, and ensure safety. Achieving these goals  requires approaches capable of making sequential, data-driven decisions in highly complex, dynamic environments. RL is particularly well-suited to  these challenges  due to its ability to learn from experience, adapt to evolving  conditions, and optimize policies for long-term objectives.  While research into prescriptive operation is still emerging, RL has already demonstrated promise in  decision support for maintenance scheduling and optimization. In these applications, RL enables data-driven strategies that improve maintenance timing, resource allocation, and system reliability \cite{knowles2010reinforcement,barde2019optimal}. Recent studies have demonstrated RL’s potential in domains such as energy systems~\cite{rokhforoz2023safe}, and manufacturing~\cite{chen2025reinforcement}, underscoring its growing relevance for PHM. A particularly promising development is the rise of Physics-Informed Reinforcement Learning (PIRL), which provides systematic approaches for integrating physical knowledge into RL \cite{banerjee2025survey}. A summary of the application contexts for RL is provided in Table \ref{checklist:RL}.

Building upon the PIRL framework, there are several concrete approaches for incorporating physical knowledge into RL systems. Physical priors can be integrated into RL using  \textbf{safety filters} by refining RL agent actions to ensure operational constraints are never violated. In PHM, such filters can prevent RL agents from selecting actions that would cause excessive equipment wear, violate health limits, or trigger unsafe states during both exploration and deployment ~\cite{rokhforoz2023safe}. 
\textbf{Physics-Informed Neural Networks} \cite{raissi2019physics} embed domain-specific knowledge, such as degradation models or conservation laws, directly into the policy or value function architecture. The Phyllis framework for data center cooling control \cite{wang2023phyllis},  which incorporates  thermodynamic equations into the RL pipeline, achieves faster and safer adaptation to changing operational conditions and improved energy efficiency compared to conventional methods.

Incorporating \textbf{task-related inductive bias} by embedding physical models or system structures within the RL environment further narrows the hypothesis space and guides agents toward physically plausible behavior. Model-based RL approaches are especially applicable to PHM, as they can integrate detailed system dynamics and degradation processes into the agent’s world model \cite{schena2024reinforcement}. This is illustrated by the reinforcement twinning framework, in which digital twins are co-trained with RL agents, and by the work of Bellani et al.~\cite{bellani2019towards}, who demonstrated that embedding a physical degradation model and RUL estimation into an RL environment enables optimal operation and maintenance policies that outperform traditional rule-based strategies by adapting proactively to real-time degradation feedback \cite{chao2022fusing,tian2022real}. Furthermore, \textbf{Graph-structured representations} offer another form of inductive bias, particularly for systems with complex inter-dependencies. For example, \cite{zheng2024combining}  modeled aircraft maintenance stand scheduling as a heterogeneous graph of maintenance stands and items, using graph neural networks and deep RL to efficiently capture system relationships. This approach led to superior scheduling performance compared to traditional heuristics, illustrating the advantages of graph-based inductive bias in complex PHM decision-making \cite{liang2025resilience}.

\begin{figure}
    \centering
    \includegraphics[width=1\linewidth]{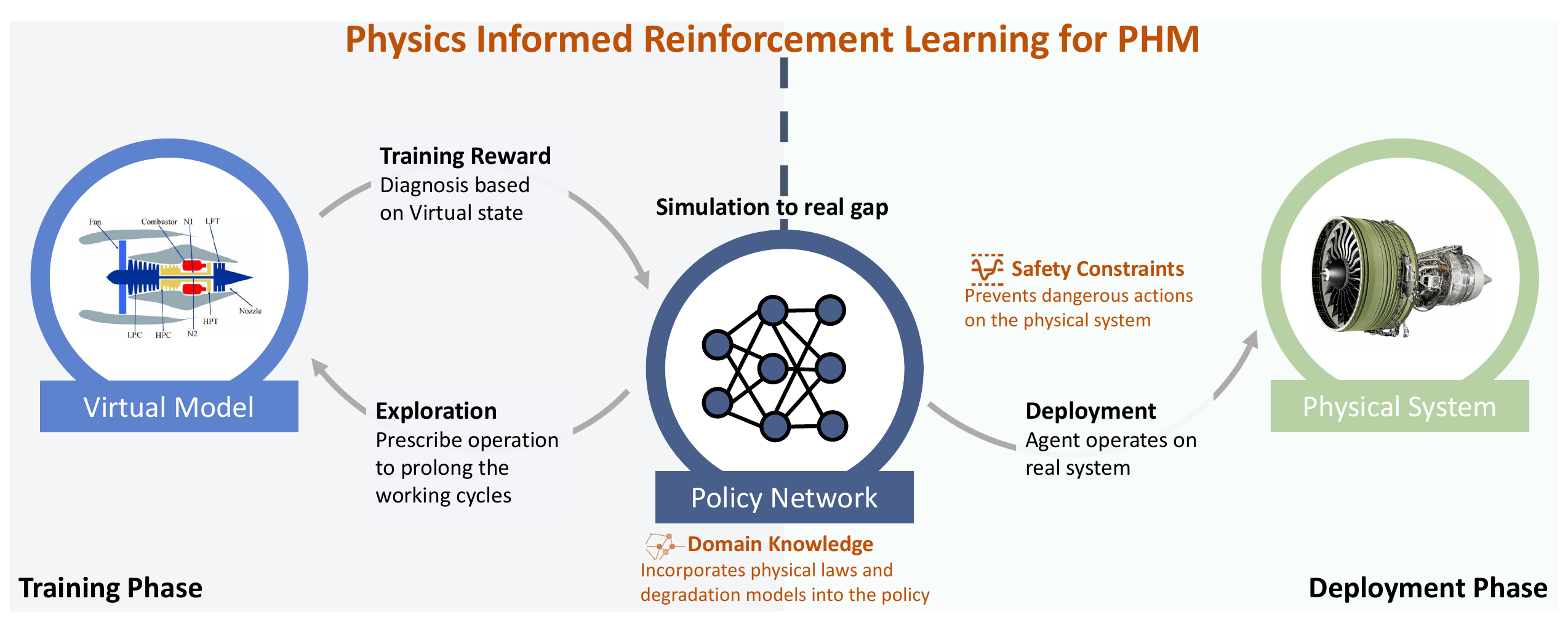}
    \caption{The framework shows safe policy training in a virtual model (left) followed by deployment on physical systems (right), with physics-informed RL including safety constraints and domain knowledge integration. The simulation-to-real gap represents a key transfer learning challenge in real-world deployment.}
    \label{fig:rl}
\end{figure}

In summary, the integration of physical knowledge and inductive biases into RL architectures marks a major step toward safer, more interpretable, and more effective PHM solutions, an overview can be seen in Figure \ref{fig:rl}. These advances are paving the way for RL-driven maintenance and operational strategies that are robust to real-world complexities and capable of delivering both predictive and prescriptive value.

\subsection*{Advantages and Limitations}

RL offers several distinct advantages over traditional rule-based or supervised learning methods. System degradation in PHM is inherently complex and uncertain, making learning-based methods especially valuable for enabling proactive, adaptive operation. RL is inherently designed for sequential decision-making, enabling agents to optimize maintenance and operational actions over time and adapt strategies dynamically as system conditions evolve. Additionally, RL also supports multi-objective optimization, balancing competing goals -- such as minimizing downtime, reducing maintenance costs, and maximizing safety or reliability --through carefully designed reward functions \cite{zhang2021model,marchetti2024deep}. 

Moreover, RL can be combined with model-based approaches and digital twins, leveraging simulation environments to train agents safely before deployment  \cite{liang2025novel}. This reduces real-world risk and enables robust sim-to-real transfer. Incorporating physics-informed representations and domain knowledge further enhances interpretability, data efficiency, and generalization, making RL highly effective for proactive PHM solutions.

However, several fundamental limitations currently constrain the widespread  deployment of RL in PHM applications. One of the most significant challenges is sample efficiency: RL agents require extensive environment interaction, often impractical or unsafe in industrial settings, necessitating high-fidelity simulators or digital twins for offline training. However, simulators must accurately capture complex system behaviors and degradation mechanisms, and even then, there is a persistent risk of a sim-to-real gap, where policies that perform well in simulation may not generalize to actual operations (as discussed in Section \ref{sim2real}). 

Another major challenge lies in the design of reward functions. Translating complex maintenance objectives—such as balancing cost, reliability, downtime, and risk—into a scalar reward signal that guides learning is inherently difficult. Poorly specified rewards may incentivize unintended or even unsafe behaviors. Achieving a balance between leveraging physics-informed or domain-specific biases and maintaining sufficient flexibility for learning is critical: overly rigid biases can hinder the agent’s ability to adapt, while insufficient guidance may lead to slow or unsafe policy development.

Finally, many RL algorithms are challenged by delayed and sparse rewards, partial observability, and the non-stationary, high-dimensional nature of PHM environments. These factors collectively limit the direct applicability of RL in safety-critical and economically sensitive maintenance contexts. Addressing these challenges remains a central research focus, requiring advances in simulation, safe exploration, reward engineering, and the integration of physical and expert knowledge within the RL framework.

\subsection*{Unexplored and Emerging Directions}

Combining Physics-Informed Machine Learning with model-based RL \cite{banerjee2023surveyphysicsinformedreinforcement} offers a promising frontier for PHM. By embedding physical laws and constraints directly into simulation environments, PIML enables the construction of more accurate and generalizable system models, a critical requirement for effective model-based RL. This synergy has already proven valuable in robotic systems, where PIML-informed dynamics models allow RL agents to learn more effective policies and generalize across a broader range of operational scenarios
\cite{ramesh2023physics}. Extending this approach to PHM could yield more robust, efficient, and trustworthy maintenance and operational strategies that adapt dynamically to the complexities of real-world assets.

Furthermore, differentiable physics simulators show great potential, as they expose analytic gradients of simulation outcomes with respect to control actions \cite{xing2025stabilizing}. Gradient information let agents optimise long‑term maintenance or operational objectives with orders‑of‑magnitude fewer samples than classic model‑free methods, even when the underlying dynamics are highly nonlinear.

Equally important is the choice of state representation. Augmenting or replacing conventional state representations with physical parameters, health indicators, and degradation metrics can make RL agents more sample-efficient and interpretable. In PHM, these physics-informed features -- often derived from domain expertise or monitoring data -- enable agents to focus on the most relevant aspects of system health and degradation. This approach not only accelerates learning and improves generalization but also enhances the transparency and trustworthiness of decision-making in practical maintenance scenarios.

Reward design, traditionally a stumbling block, is also being re‑examined. Large language models can translate free‑form operational goals or maintenance guidelines into dense, context‑aware reward signals, allowing RL agents to follow high‑level instructions expressed in natural language \cite{xie2024text2rewardrewardshapinglanguage}. When combined with the physics‑based biases above, such language‑conditioned shaping offers a practical path toward adaptive, transparent and domain‑compliant decision support for PHM.

%% file: PaperII/10_METHOD_FAST_ADAPTATION.tex
\begin{table}[htbp]
\centering
\belowrulesep=0pt
\aboverulesep=0pt
\renewcommand\arraystretch{2}
\begin{tabular}{@{}ccp{8cm}@{}}
\toprule
\multirow{2}{*}[-1ex]{Applications in PHM} &  Type of tasks         & Fault detection; Fault diagnosis; RUL prediction. \\ \cmidrule(l){2-3} 
                                     & Addressed challenges  & Fast adaptation to a novel dataset or task. \\ \midrule
\multirow{4}{*}{Requirements}        & Prior knowledge        & Consistent observational biases across small datasets. \\ \cmidrule(l){2-3} 
                                     &  Data                  & Dataset with few samples (5-100); Multiple small data sets; Unlabeled data  \\ \cmidrule(l){2-3} 
                                     & Type of bias           & Observational bias. \\ \cmidrule(l){2-3} 
                                     & Assumptions            & Observational biases can be learned by a single model, regardless of the task. This means a training task can be formulated at the meta-level, or a suitable pre-training task can be defined.  \\ \midrule
\multirow{2}{*}[-4ex]{Summary}             & Advantages             & Fast adaptation to novel tasks; Data efficiency; Robustness to distribution shifts; Integration of various data sources. \\ \cmidrule(l){2-3} 
                                     & Disadvantages          &  Computational inefficiency; Difficult task definition; Slow Adaptation in complex tasks; Limitations on task diversity\\ \bottomrule
\end{tabular}
\vspace{0.5cm}
\caption{Checklist of \textit{Fast Adaptation Methods}}
\label{checklist:fast_Adapation}
\end{table}

Physics-Informed Machine Learning (PIML) has advanced PHM by embedding physical principles into machine learning models, leading to better generalization and data efficiency, especially in safety-critical and data-sparse domains. However, real-world PHM applications often encounter rapidly changing conditions, new system configurations, and unexpected operational scenarios \textemdash that require not just robust, physics-consistent models but also the ability to adapt quickly.

Recent research has sought to address fast adaptation by developing adaptation and transfer learning mechanisms directly within PIML frameworks, enabling models to update or recalibrate in response to new data or shifts in system dynamics. For example, a recently proposed deep transfer operator learning framework enables fast and efficient learning of heterogeneous tasks despite substantial differences between source and target domains \cite{goswamiDeepTransferOperator2022}. At the same time, fast adaptation methods from meta-learning and in-context learning provide complementary capabilities for purely data-driven PHM models, allowing them to learn from limited examples, exploit contextual information, and generalize across tasks and domains with minimal manual intervention. This is particularly valuable in fleet-scale PHM, where operational conditions vary widely and where obtaining sufficient labeled data to retrain models across diverse operating scenarios is increasingly impractical at scale.

Fast adaptation methods build on knowledge acquired during an initial training phase across multiple related tasks or systems. This knowledge is embedded into the model's parameters or conditioning mechanisms and can  be efficiently reused when adapting to new, unseen systems or operating conditions. 
The unifying objective of such approaches is to leverage prior experience to enable rapid task transfer under scarce supervision and shifting operating conditions, despite differences in formal assumptions across the different approaches.
This problem is often framed within the paradigm of few-shot learning, where
a model must adapt to a new task after seeing only a few labeled examples 
\cite{songComprehensiveSurveyFewshot2023a}.
In the standard N‑way K‑shot setting, the model is required to classify inputs into N categories after observing only K labeled samples per class, with K typically ranging from one to ten \cite{wangGeneralizingFewExamples2020a}. Few‑shot learning methods can be broadly grouped into two categories: those that explicitly update model parameters for each new task, and those that avoid weight updates by relying instead on prompting or conditioning mechanisms for fast adaptation. An important special case is zero-shot learning, corresponding to the $K=0$ limit of few-shot learning. Zero-shot learning is particularly attractive for PHM applications, as it allows deployment on unseen assets, operating conditions or fault types without the need for costly data collection or retraining. Instead, it exploits prior knowledge and structural similarities between tasks to generalize across domains \cite{
zhangEffectiveZeroshotLearning2023, 
xuZeroshotFaultSemantics2023, 
chenDeepAttentionRelation2023}.

A key distinction among fast adaptation methods lies in whether they require parameter updates during adaptation or operate in a parameter-free manner.
Metric‑based few‑shot methods, such as Siamese networks \cite{bromley1993signature}, Prototypical Networks \cite{snell2017prototypical}, and Relation Networks \cite{sung2018learning}, fall into the latter category. These approaches learn an embedding space where samples from similar classes cluster closely together, enabling classification by comparing distances between embeddings rather than retraining the model. 

\begin{figure}[h]
    \centering
    \includegraphics[width=0.8\linewidth]{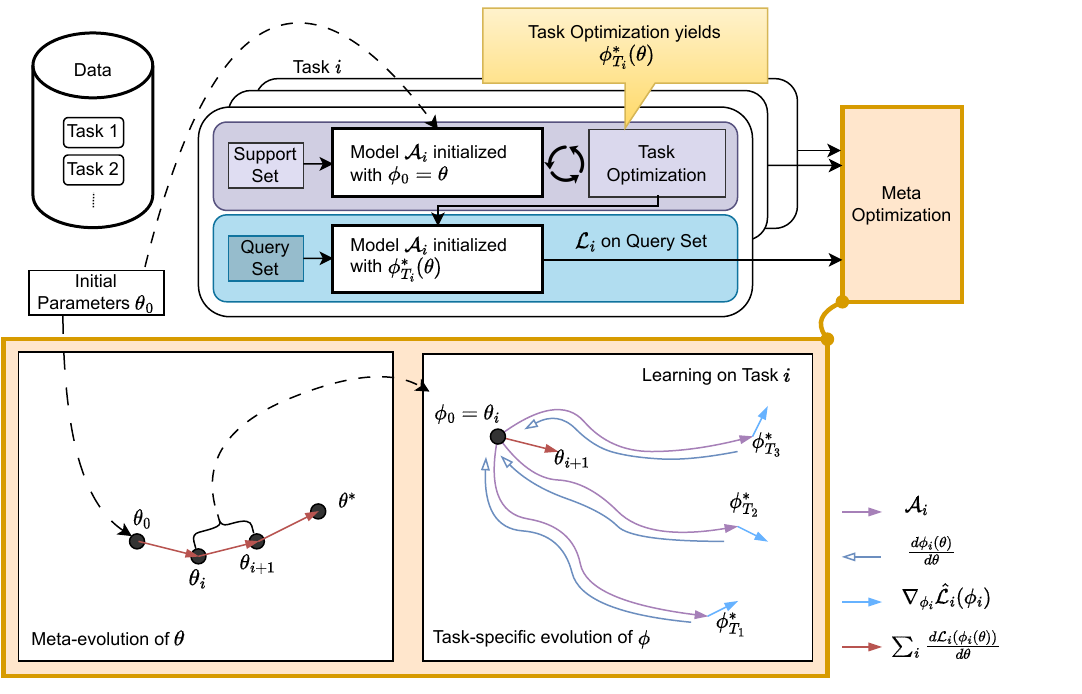}
    \caption{Model-Agnostic Meta-Learning (MAML). For each task, the model starts from shared meta-parameters $\theta_0$, adapts on a support set to obtain task-specific parameters, and then uses the query set to update $\theta$ for better generalization. The bottom-left inset shows meta-parameters converging toward the optimal initialization $\theta^*$, while the bottom-right inset illustrates task-specific adaptation trajectories.}
    \label{fig:few_shot-meta-overview}
\end{figure}

Meta‑learning, often described as "learning to learn," takes a different approach by training across a distribution of related tasks to discover a parameter initialization that is broadly useful. This initialization allows for rapid adaptation to new tasks or systems with only a few labeled examples \cite{wangGeneralizingFewExamples2020a,hospedalesMetaLearningNeuralNetworks2022, liSmallDataChallenges2024}. Among the many meta‑learning frameworks, Model‑Agnostic Meta‑Learning (MAML) \cite{Finn2017} has gained particular traction in PHM research because of its architecture independence. MAML directly optimizes model parameters to find a general initialization, defined as a point in parameter space from which gradient-based updates quickly converge to task-specific optima, as illustrated in Figure \ref{fig:few_shot-meta-overview}. The algorithm uses a two-level optimization process: an inner loop that adapts the model to individual tasks using a few examples, and an outer loop that updates the initial parameters based on how well the model performs across many different tasks. In PHM, this approach enables the model to learn a good starting point that generalizes across the fleet, so that for example, when deployed on a new asset, 
the model can quickly adapt using only limited data from that specific unit 
\cite{rajeswaranMetaLearningImplicitGradients2019, hanHybridGeneralizationNetwork2021}.

The category of parameter‑free, prompt‑conditioned fast adaptation methods keeps the model weights frozen and performs adaptation entirely at inference time \cite{dongSurveyIncontextLearning2024a}. 
Unlike few-shot learning, which typically fine-tunes model parameters using a small set of labeled examples,
in‑context learning (ICL) typically leverages large pre-trained foundation models by concatenating the support examples with the query directly in the input context. First identified in large language models \cite{brownLanguageModelsAre2020}, this "few‑shot prompting" capability has since been demonstrated across various types of foundation models, including those designed for graph-structured data and multimodal scenarios \cite{huang2023prodigy,sunGenerativeMultimodalModels2024}. In-context learning has proven effective for both classification and regression tasks using real and synthetic pre‑training data \cite{tagaTimePFNEffectiveMultivariate2025, hollmannAccuratePredictionsSmall2025}.

\subsection*{Applications in PHM}

Fast adaptation is a central objective in PHM and is often studied through the lens of few-shot learning or related meta-learning paradigms. In PHM, \textbf{few-shot learning} has been explored extensively to tackle label-scarce classification problems, particularly in fault diagnosis \cite{liSmallDataChallenges2024}.
Early applications of few-shot learning in PHM primarily built on established methods, combining existing metric-based few-shot learning methods, such as Siamese and prototypical networks, with neural architectures that were already proven effective in the field, particularly convolutional neural networks.
For example, Yang et al.~\cite{yangFewshotLearningRolling2020} demonstrate how a few-shot learning method can be applied in bearing fault diagnosis, employing Siamese convolutional neural networks 
, while Zhao et al.~\cite{zhaoIndustrialFaultDiagnosis2021} propose a convolutional prototype network for chemical process fault diagnosis.
More recent work in metric-based few-shot learning has begun to integrate transformer-inspired architectures, especially attention mechanisms. 
For instance, Hu et al.~\cite{huFewshotTransferLearning2022}
propose a relational convolutional block attention network for bearing fault diagnosis under few-shot conditions, specifically targeting the imbalance between abundant artificial damage data and limited real-world damage data.
Building on this direction, Wang et al.~\cite{wangNovelTransformerbasedFewshot2024a} proposed an enhanced transformer model that leverages the same attention mechanism but is trained with a novel asymmetric loss function to improve robustness against label noise infew-shot bearing fault diagnosis under variable speed conditions

Complementing these architecture-focused advances, recent research has also revisited the design of distance measures for metric few-shot learning. 
Costa et al.~\cite{costaFewshotGenerativeCompression2025}, for instance, employ a pre-trained variational autoencoder and introduced a novel Kolmogorov-complexity-based distance in the latent space, enabling system health estimation with minimal labeled data
based on most similar k-neighbors.

In parallel, other works have shifted focus from distance metrics to the learning paradigm itself, investigating how to decouple feature extraction from episodic training and integrate meta-learning with multi task learning. 
Wang et al.~\cite{wangMetricbasedMetalearningModel2021} 
address this by first training a feature extractor through supervised classification on source-domain labels, then freezing it and performing episodic metric learning in the resulting feature space. This two-stage process improves knowledge transfer and yields higher accuracy for bearing and gearbox fault diagnosis under limited data conditions.
Building on this idea, 
Lei et al.~\cite{leiPriorKnowledgeembeddedMetatransfer2023} demonstrate that feature extraction can also be enhanced through semantic self-supervised pretraining. They propose a prior knowledge-embedded meta-transfer learning framework that jointly optimizes multi-task metric meta-learning and adaptive information fusion, achieving robust performance in bearing fault diagnosis under scarce-label settings.

While metric-based few-shot learning primarily targets classification tasks, \textbf{meta-learning} offers a more versatile framework for fast adaptation in PHM \cite{zhangIntelligentFaultDiagnosis2022}.
Due to its flexibility and model-agnostic nature, MAML has become one of the most widely adopted meta-learning approaches in the PHM literature. Applications span from the health monitoring of rotary machine components, such as bearings, to machining and manufacturing processes, as well as turbofan engine prognostics.
For example, Ding et al.~\cite{dingMetaDeepLearning2021} integrate MAML into a meta learning framework 
for bearing prognostics and fault diagnostics, while Yang et al.~\cite{yangNovelCrossdomainFault2022} 
employ MAML with adaptive threshold networks to improve cross-domain fault diagnosis.

Building on these classification-oriented studies, large body of work extends MAML to regression tasks, most notably for remaining useful life estimation in turbofan engines 
\cite{
moFewshotRULEstimation2023, 
yangFewshotRemainingUseful2024, 
wangDistributionAgnosticProbabilisticFewShot2024},
as well as for tool wear prediction in machining processes
\cite{liNovelMethodAccurately2019}.

Recently, several extensions of MAML have been proposed to better address the specific challenges of PHM tasks.
For example, Wang et al.\cite{wangDistributionAgnosticProbabilisticFewShot2024} extend MAML to a Bayesian framework, enabling probabilistic modeling of multiple failure modes and more reliable estimation of remaining useful life \cite{wangDistributionAgnosticProbabilisticFewShot2024}.
Building on this line of work, Yang et al.~\cite{yangMetalearningDeepFlow2024} integrate MAML with conditional normalized flow to infer richer posterior distributions in the kernel features, thereby extracting more discriminative information from engine and bearing degradation data.
In a related direction, Ding et al.~\cite{dingMechatronicsEquipmentPerformance2022} 
combine MAML with pseudo labeling to enable few-shot prognostics when large amounts of unlabeled historical data are available.

Beyond these extensions of MAML, meta-learning has also been integrated with GNNs for remaining useful life prediction under variable operating conditions by constructing a spatio-temporal graph that integrates structural data with multi-source, multi-channel signal, demonstrating the applicability of meta-learning to non-euclidean domains
\cite{dingGraphStructureFewshot2024}.
In addition to graph-based approaches, physics-informed meta learning has also been explored. 
Li et al.~\cite{liPhysicsinformedMetaLearning2022}
propose a framework that integrates empirical tool wear models into data-driven modeling, allowing the meta-learning process to adapt quickly across wear stages while remaining consistent with physical laws. Similarly, Majumdar et al.~\cite{majumdarHxPINNHypernetworkbasedPhysicsinformed2024} propose a hypernetwork-based physics-informed neural network model 
for real-time monitoring of an industrial heat exchanger.
While a large body of literature has focused on MAML-based strategies, recent results 
demonstrate that metric-based few-shot approaches remain competitive, and in fact outperform MAML 
on two bearing datasets using recent transformer architecture
\cite{wangNovelTransformerbasedFewshot2024a}.

Zero-shot methodology has recently gained traction in PHM as a way to detect previously unseen fault types without requiring labeled training data. Approaches in this direction span several complementary strategies. Some integrate physical knowledge with generative models, embedding physics-based priors for zero-sample diagnosis \cite{muIntegratingPhysicalKnowledge2025}. Others leverage semantic descriptions of faults, where bi-linear compatibility functions map unseen to seen classes \cite{zhangEffectiveZeroshotLearning2023}.

Building on semantic spaces, Ma et al.~\cite{maBroadZeroshotDiagnosis2024} diagnose compound faults from single-fault training data by projecting vibration features into a descriptive semantic space. Chen et al.~\cite{chenPyramidtypeZeroshotLearning2023} exploit information granularity, predicting pyramid-structured attributes and matching them to expert-defined descriptions. Xu et al.~\cite{xuZeroshotFaultSemantics2023} cluster representations in semantic space to infer category centroids for unseen fault detection, while Cai et al.~\cite{caiGeneralizedZerosampleIndustrial2025} combine out-of-distribution detection with clustering and semantic attribute projections to refine classification of both seen and unseen faults.

Although no dominant paradigm has yet emerged, this growing body of work demonstrates the promise of zero-shot learning for generalizing fault diagnosis to entirely new failure modes.

In parallel to few-shot and meta-learning, \textbf{in-context learning} has so far seen limited adaptation in PHM, though it shows promise in related anomaly detection tasks. For instance, Zhu and Pang~\cite{zhu2024toward} proposed an in-context residual learning framework that detects anomalies across diverse datasets, spanning industrial, medical, and semantic datasets, without requiring additional training, demonstrating the adaptability of ICL to heterogeneous settings. Building on this idea, Gu et al.~\cite{gu2024anomalygpt} employed a large vision-language model to generate textual anomaly scores in few-shot in-context scenarios, aiming to enhance transparency and interpretability in industrial anomaly detection. These early efforts suggest that ICL could offer a new paradigm for PHM by enabling rapid, training-free adaptation with improved interpretability.

\subsection*{Advantages and Limitations}

Parameter-updating methods such as MAML learn parameter initializations that capture shared patterns across operating conditions or fleet units, enabling rapid adaptation to new configurations or assets with minimal labeled data and few gradient steps \cite{wang2020few}. This significantly reduces both calibration time and the reliance on extensive fault libraries for each unit. However, the bi-level optimization introduces substantial computational overhead from backpropagating through inner and outer loops, which can be prohibitive in industrial applications \cite{finn2017model,vettoruzzoAdvancesChallengesMetaLearning2023}. 

In addition, MAML performance remains dependent on the underlying model architecture. 
These methods are also sensitive to domain shift, when deployment conditions differ from those seen during meta-training, the learned initialization may adapt poorly or produce miscalibrated predictions \cite{she2025meta}. Additionally, changes in sensor modalities or data structures typically require retraining, since the meta-learned parameters are architecture-dependent \cite{songComprehensiveSurveyFewshot2023}. 
Compared to MAML, metric-based few-shot methods have the advantage of avoiding inner-loop gradient updates at test time, which makes them faster, more stable, and less sensitive to optimization hyperparameters, as inference typically reduces to embedding and nearest-neighbor comparison rather than meta-optimization. Nevertheless, these methods can struggle to generalize when new classes or tasks significantly differ in distribution from the training data, since the learned metric or embedding space may fail to capture meaningful similarities and the resulting metric can become unreliable in unseen domains.

Parameter-free approaches like in-context learning enable adaptation by providing a few labeled examples directly within the input, allowing the model to infer task patterns without any parameter updates. This offers significant flexibility since the same model can handle diverse tasks and data structures simply by changing the provided examples. However, these methods rely on attention mechanisms that scale quadratically with sequence length, creating substantial computational overhead that limits both the length of time series that can be processed \cite{rasul2024vq} and the number of in-context examples that can be provided. Performance depends heavily on carefully selecting, ordering, and curating the in-context examples \cite{liWhyDoesIncontext2024,nejjar2024context}.

\subsection*{Unexplored and Emerging Applications in PHM}

Building on recent advances in meta-learning advances opens up several promising research directions for PHM. Meta-reinforcement learning, for instance, shows how agents can acquire learning priors that enable rapid adaptation to new tasks from few interactions \cite{beck2025tutorial}. Applied to PHM, this could yield control policies that adjust on the fly to new machines, fault modes, or operating conditions, alleviating the need to retrain from scratch for each scenario and reducing the computational burden of RL applications (Section \ref{RL}).

Complementing this, physics-informed meta-learning shows strong potential for generating surrogate models that generalize across PDE-driven tasks and adapt to new system parameters with minimal data \cite{cho2023hypernetwork, bihlo2024improving}. For PHM, such approaches could enable fast-calibrating digital twins that adapt to new assets with limited sensor data while respecting known degradation physics.

A third direction is unsupervised meta-learning, which generates diverse training tasks from unlabeled data through augmentations and mixing strategies to train models for transferable adaptation patterns \cite{vettoruzzo2025unsupervised}. Unlike foundation models where in-context learning emerges incidentally from general pretraining, this approach explicitly optimizes for task adaptation. In PHM, this could allow models to learn adaptation strategies directly from abundant unlabeled operational data and later specialize to new fault modes or asset configurations without extensive labeled fault libraries.

Finally, foundation models present a complementary approach. General-purpose language models have been adapted for equipment monitoring \cite{liuSurveyFoundationModels2024}, with specialized variants developed for mechanical fault diagnosis \cite{heFewshotLearningPlastic2023}. Recent advances show that time-series foundation models can be reprogrammed by keeping language-model backbones frozen while adding lightweight adapters for cross-domain transfer \cite{jin2024timellm}, or by treating series as text tokens to enable zero-shot forecasting \cite{gruver2023large}; unified architectures further handle forecasting, classification, and anomaly detection within a single backbone \cite{gao2024units}, with multimodal grounding and federated variants improving instruction-following over visualized traces and privacy-preserving cross-domain transfer \cite{yoon2024my,liu2024time}. These capabilities have been successfully transferred to fault detection applications \cite{jiangZeroShotFaultDiagnosis2025,10592003}. 

%% file: PaperII/11_Generalization.tex
\subsection*{Introduction to Domain Generalization} \label{dg}

\begin{table}[t]
\centering
\belowrulesep=0pt
\aboverulesep=0pt
\renewcommand\arraystretch{2}
\begin{tabular}{@{}ccp{8cm}@{}}
\toprule
\multirow{2}{*}[-1ex]{Applications in PHM} &  Type of tasks         & Fault detection; RUL prediction. \\ \cmidrule(l){2-3} 
                                     & Addressed challenges  & Poor generalization  to different machines or conditions. \\ \midrule
\multirow{4}{*}{Requirements}        & Prior knowledge        &  Frequency components in  neural networks  and  Fourier-based analysis.  \\ \cmidrule(l){2-3} 
                                     &  Data                  & Labeled source domain data.             \\ \cmidrule(l){2-3} 
                                     & Type of bias           & Frequency domain prior of data and neural networks. \\ \cmidrule(l){2-3} 
                                     & Assumptions            & The labeled source domain data is available while the target domain data is unavailable.    \\ \midrule
\multirow{2}{*}[-4ex]{Summary}             & Advantages             & Improved generalizability with incorporation of Frequency domain information. \\ \cmidrule(l){2-3} 
                                     & Disadvantages          &  Difficulty of capturing all possible domain variations and the risk of overfitting to source domains. \\ \bottomrule
\end{tabular}
\vspace{0.5cm}
\caption{Checklist of \textit{Prior Knowledge-based Domain Generalization}}
\label{checklist:dg}
\end{table}


A fundamental challenge in PHM is domain shift, which occurs when the data distribution in the operational environment differs from the training dataset. This distributional discrepancy arises from several factors, including variations in operating and environmental parameters \cite{nejjar_domain_2024} such as load, speed, and temperature, and humidity, as well as differences in system configurations such as sensor types and placements, calibration settings, or overall system design. Furthermore, temporal factors like system degradation and seasonal changes can introduce gradual data shifts~\cite{xiao2024feature}. Consequently, this mismatch between source and target domains often leads to a significant deterioration in model performance, limiting the generalizability of machine learning solutions to new conditions \cite{Wang2019,nejjar2024uncertainty}. 

\begin{figure*}[t]
  \centering
  \includegraphics[width=0.65\linewidth]{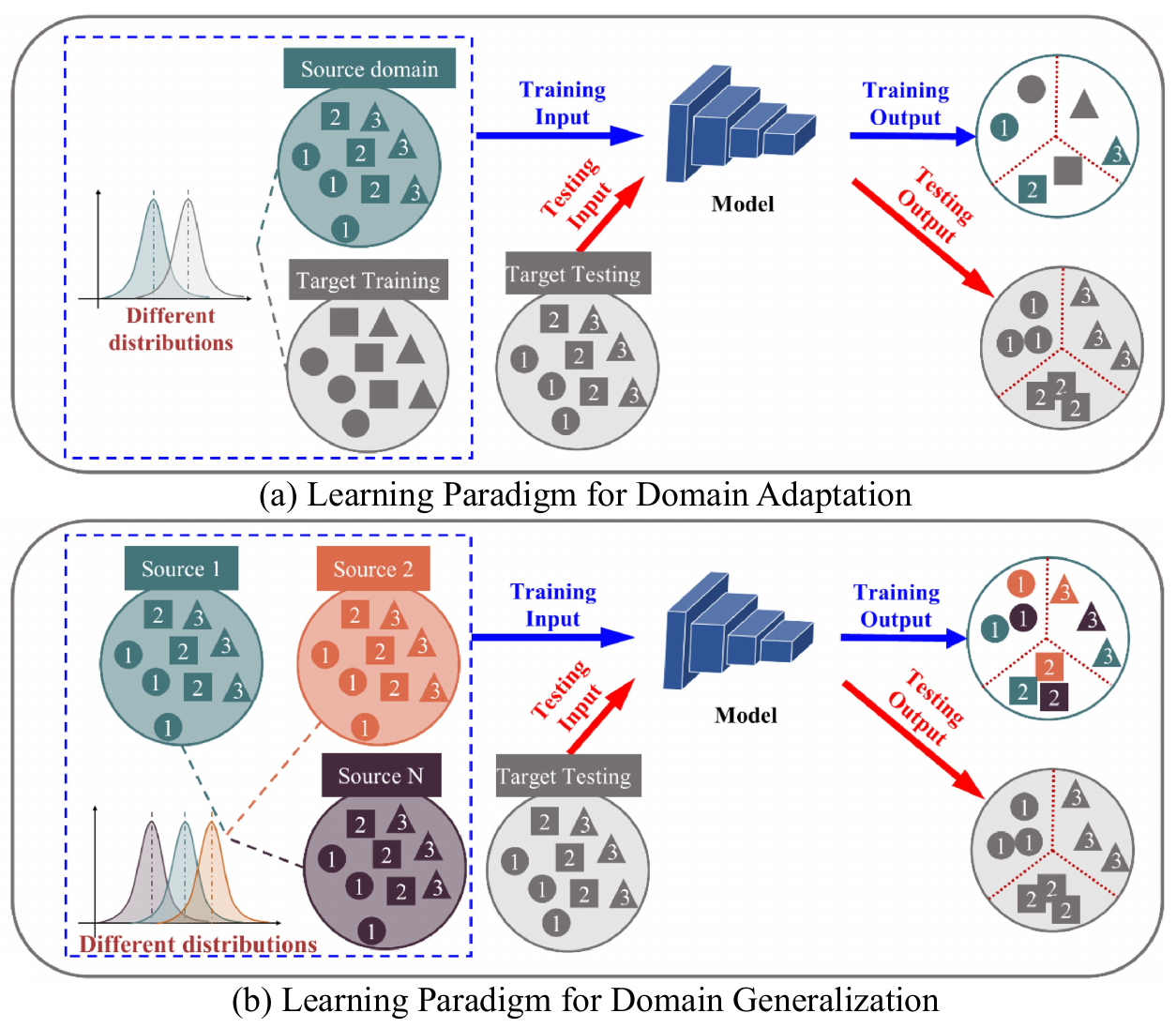}
   \vspace{-0.3cm}
   \caption{Differences between learning paradigms for domain adaptation and domain generalization~\cite{zhao2024domain}.}
   \label{fig:frame}
\end{figure*}

To address the challenge of domain shifts, several learning paradigms have been explored. While fast adaptation methods (Section \ref{sec:few_shot}) such as few-shot learning~\cite{bhatt_weakly-supervised_2023}, meta-learning~\cite{ding2020}, in-context learning~\cite{akyurek2022learning}, and domain adaptation (DA)~\cite{nejjar2024uncertainty} are effective, they typically rely on the availability of samples from the target domain. However, this assumption may not hold in many real-world scenarios, such as fault diagnosis and remaining useful life prediction. 
Domain generalization (DG)~\cite{9782500,dong2023SimMMDG}, in contrast, offers a more promising approach. By leveraging knowledge from several source domains, DG techniques aim to develop models that exhibit improved generalization capabilities on target domains not encountered during training. An illustration of the differences between domain adaptation and generalization is shown in Figure \ref{fig:frame}.

DG can be generally categorized into three main approaches~\cite{9782500}: data manipulation, representation learning, and learning strategies. Data manipulation techniques enhance generalization by increasing the diversity of training data through augmentation~\cite{Zhou_Yang_Hospedales_Xiang_2020,nejjar2024sfosda}. Representation learning methods seek to acquire domain-invariant representations using domain-adversarial neural networks~\cite{ganin2016domain,li2018domain}, explicit feature distribution alignment~\cite{tzeng2014deep}, and instance normalization~\cite{pan2018two}. Additionally, learning strategies such as meta-learning~\cite{li2018learning}, gradient operation~\cite{huang2020self}, and self-supervised learning~\cite{dong2024moosa} have been explored to improve generalization performance. For example, \cite{huang2020self} forces the network to activate features that correlate with labels, and \cite{dong2024moosa} designs multimodal self-supervised pretext tasks such as masked cross-modal translation and multimodal Jigsaw puzzles to improve generalization.

\subsection*{Prior Knowledge-based Domain Generalization}

Beyond purely data-driven strategies, domain generalization can also benefit from incorporating prior knowledge that captures invariances shared across domains. Such priors constrain the learning process toward features that are less sensitive to distributional shifts, thereby improving robustness and transferability.

Variations in operating conditions, environmental factors, and system configurations described above can alter the statistical properties of the measured signals, often affecting their frequency content. Since certain frequency components remain relatively stable across domains, they can serve as robust priors for domain generalization.
Leveraging such frequency-domain priors allows models to focus on features that are invariant to domain shifts. 
For example, the Fourier phase of an image, which encodes high-level semantic structures, is robust to common domain shifts~\cite{xu2021fourier}.
Similarly, deep neural networks exhibit an inherent bias towards learning certain frequency components during training~\cite{lin2023deep}. 
Incorporating these frequency-based priors into the learning objective can guide models toward more transferable features, thereby improving robustness and generalization.

A common technique involves manipulating the Fourier components directly, where phase information (structural content) is preserved while the amplitude information (domain-specific characteristics) is swapped with that of other domains~\cite{xu2021fourier}. This strategy retains core structural information while mitigating domain-specific variations. To further refine this, some methods introduce a soft-thresholding function to remove small, potentially noisy frequency components that might be domain-specific~\cite{li2024takes}. 

Beyond static Fourier manipulation, recent methods combine frequency priors with physically grounded augmentation strategies.  For instance \cite{Chattopadhyay_2023_ICCV} perturbs the Fourier amplitude in proportion to the original energy distribution, ensuring that the perturbations remain physically plausible while introducing controlled variability. This targeted spectral modification exposes the model to a broader range of frequency configurations, encouraging the extraction of structural cues that remain stable across domains. Similarly, PhysAug \cite{xu2025physaug} integrates physical guidance with frequency-domain augmentation for single-domain generalized object detection. By combining physically motivated perturbations with spectral manipulations, PhysAug improves model robustness without introducing unrealistic artifacts. 

In addition to spectral invariance, prior knowledge can also take the form of structural and causal relationships derived from the underlying physics of the system. 
For instance ~\cite{HUANG2025130187} encodes causal dependencies between physical variables and measurement signals, integrating them into the learning process via physics-based constraints. By constraining the learned features to adhere to known physical mechanisms, the model is encouraged to focus on domain-invariant relationships, improving generalization to operating conditions never encountered during training.

Finally, the inherent frequency bias of deep neural networks can itself be leveraged or corrected. Such models tend to learn lower-frequency components first, which can create a dependency on "frequency shortcuts" patterns that are highly correlated with the source domain but do not generalize well ~\cite{xu2019training,xu2018understanding}. Recognizing this, Lin \textit{et al.}~\cite{lin2023deep} introduced deep frequency filtering, which modulates frequencies in the latent space. This method uses an attention mechanism in the frequency domain to amplify transferable features and suppress non-transferable ones.

\subsection*{Application of Domain Generalization in PHM}

In many practical PHM applications, obtaining representative labeled data from all possible future operating conditions or equipment configurations is either prohibitively expensive or simply not feasible. In such cases, DG becomes a more critical and relevant task than DA, as it does not rely on access to target domain data during the training phase. In PHM, the successful application of DG enhances the reliability of key predictive tasks, such as RUL estimation and fault diagnosis, across diverse and previously unseen equipment. This improved predictive accuracy directly enables more efficient resource allocation. For example, by facilitating just-in-time maintenance scheduling based on reliable RUL predictions, organizations can avoid both premature component replacement and costly unexpected failures. This targeted approach optimizes the deployment of maintenance personnel and spare parts, ultimately maximizing equipment availability and reducing operational costs~\cite{su2023machine}.
The following sections review how DG approaches have been applied to the two labeled PHM tasks: fault diagnosis and remaining useful life prediction.

Domain generalization plays a crucial role in  building fault diagnostics and prognostic models that are robust to unseen operating conditions. Key strategies include data manipulation, representation learning, and advanced learning strategies~\cite{chen2025applications}. Data manipulation techniques enhance training data diversity~\cite{guo2024domain}; for instance, the domain augmentation generalization network employs multi-source augmentation and adversarial training to generate varied data, boosting the model's robustness to out-of-distribution samples~\cite{shi2023domain}. Representation learning aims to extract domain-invariant features~\cite{chu2023gear,hong2023multi}. A common strategy is to combine Domain Adversarial Neural Network for implicit feature alignment with explicit Maximum Mean Discrepancy minimization to reduce distribution gaps, an approach that has been successfully applied to diagnose rock drill faults under variable conditions~\cite{kim2023domain}. Finally, learning strategies like meta-learning directly optimize for generalization. The Meta-GENE framework~\cite{ren2023meta}, for example, uses a model-agnostic meta-learning procedure with gradient alignment to learn a domain-invariant optimization policy, achieving strong performance on gearbox fault diagnosis tasks.

Applying domain generalization to RUL prediction is essential for creating prognostic models that can adapt to new assets or changing operational profiles, a task where target run-to-failure data is often not available~\cite{da2020remaining}. RUL prediction is a regression task and is more difficult than classification because it requires precise, continuous predictions that are highly sensitive to distribution shifts. Additionally, small changes in input can lead to large output errors, making robust generalization across domains more challenging than in classification.
Data manipulation strategies generate synthetic degradation trajectories to train more generalizable regressors~\cite{xu2024asdinnorm}. The Adversarial Out-Domain Augmentation~\cite{ding2023domain} framework exemplifies this by training a generator to produce diverse pseudo-domains of degradation signals, which are then used to train a robust RUL predictor. Representation learning methods focus on creating a domain-agnostic health index~\cite{xiao2025contrastive}. This is often achieved by integrating recurrent architectures like Bidirectional Gated Recurrent Units (BGRU)~\cite{cheng2021data} with a DANN, where the BGRU captures temporal degradation patterns and the DANN aligns feature distributions across different working conditions. Advanced learning strategies are also employed, particularly for small-data scenarios. For instance, a meta-domain generalization method for tool wear prediction~\cite{wang2022meta} uses a phased approach of data splitting, meta-pre-training, and fine-tuning to effectively generalize from limited source data. 
A summary of the application contexts for domain generalization is provided in Table \ref{checklist:dg}.

\subsection*{Advantages and limitations of Domain Generalization in PHM}

Domain generalization offers significant advantages for the practical application of PHM in industrial settings. A primary benefit is the reduced dependency on labeled data from every potential target domain. In many PHM applications, acquiring comprehensive labeled datasets that encompass all operating conditions, equipment variations, and environmental factors is often prohibitively expensive, time-consuming, or impractical~\cite{zhao2024domain}. DG mitigates this challenge by enabling models trained on a limited set of source domains to generalize effectively to new, unseen target domains, thereby reducing the need for target-specific labels. Additionally, DG does not rely on access to target domain samples, thereby addressing a key limitation of DA methods.

Moreover, models developed with DG techniques demonstrate enhanced robustness and reliability when deployed in real-world environments characterized by distribution shifts. By design, these models are more adept at managing the inherent variability of operational data, resulting in more consistent and accurate predictions of faults and RUL across diverse conditions \cite{yan2024transfer}. This enhanced reliability is essential for building trust in PHM systems and supporting their adoption in safety-critical industrial and infrastructure applications.

Despite these benefits, DG faces notable limitations. It fundamentally relies on the assumption that the source domains capture enough variability to represent potential target domains. When novel or unforeseen operational shifts occur in the target environment, they may lie outside the scope of the source data, leading to degraded performance. In addition,  strong generalization typically requires a diverse and carefully curated training dataset, which can be difficult to obtain in many industrial PHM settings.

\subsection*{Unexplored or Emerging Applications of Domain Generalization in PHM}
In PHM, domain generalization is an emerging approach to enhance fault diagnostics and prognostics across diverse operational conditions. However, most existing DG research in PHM primarily focuses on unimodal data, while the real world is multimodal. Effectively integrating information from multiple modalities is crucial for improving diagnostics and prognostics performance~\cite{dong2025mmdasurvey,dong2025aeo}. One major barrier to multimodal DG in PHM is the lack of public multimodal datasets and benchmarks. For example, the accelerometer data is the predominant modality in fault diagnosis, while other modalities may also help, such as images and audio data, as motivated in Section \ref{sec:data_fusion}.  
To advance research in this area, it is essential to identify use cases where multiple modalities are critical and develop standardized benchmarks tailored to those applications. For example, in the railway monitoring system, vision modality is a good complement to time-series modality for arcing detection. An additional limitation is that most existing DG approaches for PHM operate under the closed-set assumption, which presumes identical label spaces across domains. However, in real-world applications, target domains often contain unknown faults, making it essential to detect and handle them effectively. Techniques developed from the out-of-distribution detection~\cite{li2024dpu,dong2024multiood} and anomaly detection~\cite{nejjar2022dg} domains could be leveraged and adapted to facilitate unknown class detection. Finally, exploring DG techniques for fault detection is also promising, as it addresses a fully unsupervised and considerably more challenging setting.

%% file: PaperII/13_conclusions.tex
Prognostics and Health Management has evolved into a critical discipline for ensuring the safety, efficiency, and resilience of complex engineered systems. Modern PHM extends well beyond traditional tasks such as fault detection, diagnostics and prognostics to encompass system health management, health-aware decision-making, real-time control, and fleet-level asset operation and maintenance optimization. Meeting these demands requires algorithms that are accurate and scalable, interpretable, robust to uncertainty, and consistent with physical system behavior.

While machine learning has enabled major advances in PHM by modeling complex dynamics and integrating multimodal data, purely data-driven models often 
struggle under distribution shifts, perform poorly when data is limited or noisy, and often lack physical consistency. These limitations are particularly critical in domains where degradation processes are gradual, only partially observable, and influenced by external factors not directly measured. To address these challenges, Physics-Informed Machine Learning  has emerged as a promising paradigm that integrates domain knowledge and physical constraints into the learning process, improving model reliability and trustworthiness.

In this second part of our review on Physics-Informed Machine Learning  for PHM, we have focused on the remaining two biases that complement the inductive biases reviewed in Part I: \textbf{learning biases} and \textbf{observational biases}. These two biases shape how machine learning models are trained and exposed to data in ways that embed physical knowledge, improve robustness, and enhance trustworthiness, especially in industrial settings where data is sparse, labels are limited, and degradation evolves under complex, partially observable conditions.

We examined how learning biases guide the optimization process by embedding physical principles into loss functions, training objectives, or regularization schemes. Physics-Informed Neural Networks and degradation-informed learning methods serve as prototypical examples, ensuring that predictions are consistent with known dynamics or general physical properties of degradation processes such as monotonicity and irreversibility. In parallel, observational biases enhance model robustness by curating, enriching, or reconstructing datasets that more faithfully represent system behavior. Techniques such as virtual sensing, overcoming the simulation to real domain gap of physics-based simulations, physics-informed data augmentation, and multimodal fusion help bridge gaps in observability, enabling models to generalize beyond narrowly defined training distributions.

We also highlighted how to transition from prediction to action in PHM by leveraging reinforcement learning and health-aware control, enabling intelligent decision-making grounded in physical understanding. In this context, RL closes the loop from physics to machine learning and back to the physical world, where model-driven decisions directly affect system behavior. This shift transforms PHM from a passive monitoring task into an interactive, decision-oriented discipline, where learned policies adaptively manage reliability, risk, and performance in dynamic environments. 
Despite this progress, several open challenges remain. A major limitation, as already outlined in Part I, is the lack of systematic benchmarking and validation in realistic industrial contexts. Methods are typically evaluated on synthetic or clean and controlled laboratory datasets, making it difficult to assess their robustness, transferability.
Degradation-informed approaches provide valuable flexibility, but their simplifying assumptions remain limiting, particularly in disentangling degradation from operational dynamics. Observational bias techniques such as leveraging simulated data or deploying virtual sensors carry inherent limitations. Synthetic trajectories generated from physics-based simulations often fail to reflect the complexity and variability of real-world degradation, hence leading to a simulation-to-real gap. Virtual sensing strategies that infer hidden system states from indirect sensor measurements introduce modeling and identifiability uncertainties, as sensor drift can lead to error accumulation that propagates misrepresentations of system health.

Looking ahead, several promising research directions emerge. One is the development of adaptive training strategies that adjust the influence of physical constraints based on model confidence or data coverage. Another is advancing physics-aware generative models that synthesize realistic and physically plausible degradation trajectories. In RL, there is increasing potential in embedding physical structure at different levels, including the reward function, policy architectures, world models, and value functions, enabling physics-informed and sample-efficient decision-making. Hybrid frameworks that combine partial physical knowledge, data-driven inference, and symbolic reasoning are particularly promising for scenarios where physics is implicit or only partially known. Finally, scaling to fleet-level deployment will require generalization across diverse systems and operating conditions, motivating new research in meta-learning, in-context adaptation, and domain generalization under physical constraints.

In summary, this second part of our review expands the PIML framework to include not only inductive biases but also learning and observational biases, thus completing the picture of how physical knowledge can be infused into data-driven PHM systems. By embedding physics throughout the modeling pipeline (from data to training, to decision-making) PIML offers a powerful paradigm for building scalable, reliable, and physically consistent PHM solutions that meet the demands of real-world deployment in complex, safety-critical systems.